\documentclass{article}

 \usepackage[preprint]{neurips_2026}

\usepackage[utf8]{inputenc} 
\usepackage[T1]{fontenc}    
\usepackage{hyperref}       
\usepackage{url}            
\usepackage{booktabs}       
\usepackage{amsfonts}       
\usepackage{nicefrac}       
\usepackage{microtype}      
\usepackage{xcolor}         
\usepackage{enumitem}
\usepackage{amsmath}   
\usepackage{amssymb}   
\usepackage{amsfonts}  
\usepackage{graphicx}
\usepackage{multirow}
\usepackage[table]{xcolor} 
\usepackage{colortbl}      
\usepackage{arydshln}
\usepackage{booktabs}
\usepackage{float}
\usepackage{tabularx}
\usepackage{makecell}
\usepackage[most]{tcolorbox}
\usepackage{xcolor}
\usepackage{nicematrix}

\definecolor{GPTBlock}{RGB}{248,251,255}
\definecolor{QwenBlock}{RGB}{255,250,245}
\definecolor{GPTHead}{RGB}{235,243,255}
\definecolor{QwenHead}{RGB}{255,239,222}
\newcommand{\secref}[1]{\S\ref{#1}}
\newcommand{\best}[1]{\textbf{#1}}
\newcommand{\second}[1]{\underline{#1}}
\usepackage{pifont}

\newcommand{\cmark}{\ding{51}}%
\newcommand{\xmark}{\ding{55}}%
\usepackage{caption}
\usepackage[table,dvipsnames]{xcolor}
\newcommand{\gain}[1]{\ensuremath{_{\scriptscriptstyle\textcolor{ForestGreen}{\uparrow #1\%}}}}
\newcommand{\dropc}[1]{\ensuremath{_{\scriptscriptstyle\textcolor{BrickRed}{\downarrow #1\%}}}}
\newcommand{\nochange}{\ensuremath{_{\scriptscriptstyle\textcolor{gray}{\rightarrow 0.00\%}}}}
\newcommand{\modelwithlogo}[2]{%
  \raisebox{-0.15em}{\includegraphics[height=1.1em]{figs/#1.pdf}}~#2%
}

\usepackage{xurl} 

\newcommand{\dimavailability}{%
\begin{center}
\small
\begin{tabular}{@{}l@{}}

\raisebox{-0.18em}{\includegraphics[height=1.05em]{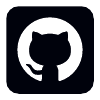}}%
\hspace{0.35em}%
\textbf{Github:}~
\href{https://github.com/ChowRunFa/DimMem}%
{\texttt{github.com/ChowRunFa/DimMem}}
\\[-0.05em]

\raisebox{-0.18em}{\includegraphics[height=1.05em]{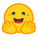}}%
\hspace{0.35em}%
\textbf{Data:}~
\href{https://huggingface.co/datasets/wtqiu/DimMem-SFT-GPT-5.4}%
{\texttt{huggingface.co/datasets/wtqiu/DimMem-SFT}}
\\[-0.05em]

\raisebox{-0.18em}{\includegraphics[height=1.05em]{figs/huggingface.pdf}}%
\hspace{0.35em}%
\textbf{Models:}~
\href{https://huggingface.co/wtqiu/DimMem-4B-Locomo}%
{\texttt{huggingface.co/wtqiu/DimMem-4B-LoCoMo}}%
\hspace{0.6em}$\vert$\hspace{0.6em}%
\href{https://huggingface.co/wtqiu/DimMem-4B-Longmemeval}%
{\texttt{DimMem-4B-LME}}

\end{tabular}
\end{center}
}

\title{DimMem: Dimensional Structuring for Efficient Long-Term Agent Memory}

\author{%
\begin{tabular}{c}
\textbf{Wentao Qiu}$^{1,2}$ \quad
\textbf{Haotian Hu}$^{1}$\thanks{Corresponding authors. Fanyi Wang is the project leader.} \quad
\textbf{Fanyi Wang}$^{1}$\footnotemark[1] \quad
\textbf{Jinwei Kong}$^{1,3}$ \quad
\textbf{Yu Zhang}$^{1}$ \\
\\[-0.5ex]
$^{1}$StepOS \quad
$^{2}$Xiamen University \quad
$^{3}$ShanghaiTech University \\
\texttt{wtqiu@stu.xmu.edu.cn} \\
\texttt{\{huhaotian,wangfanyi\}@stepos.com}
\end{tabular}
}
\begin{document}

\maketitle

\begin{abstract}
Large language model (LLM) agents require long-term memory to leverage information from past interactions. However, existing memory systems often face a fidelity--efficiency trade-off: raw dialogue histories are expensive, while flat facts or summaries may discard the structure needed for precise recall. We propose \textbf{DimMem}, a lightweight dimensional memory framework that represents each memory as an atomic, typed, and self-contained unit with explicit fields such as time, location, reason, purpose, and keywords. This representation exposes the structure needed for dimension-aware retrieval, memory update, and selective assistant-context recall without storing full histories in the model context. Across LoCoMo-10 and LongMemEval-S, DimMem achieves \textbf{81.43\%} and \textbf{78.20\%} overall accuracy, respectively, outperforming existing lightweight memory systems while reducing LoCoMo per-query token cost by \textbf{24\%}. We further show that dimensional memory extraction is learnable by compact models: after fine-tuning on the DimMem schema, a Qwen3-4B extractor surpasses LightMem with GPT-4.1-mini on both benchmarks and reaches performance comparable to, or better than, much larger extractors in key settings. These results suggest that explicit dimensional structuring is an effective and efficient foundation for long-term memory in LLM agents.
\dimavailability
\end{abstract}

\section{Introduction}
\label{sec:introduction}

Large language model (LLM) agents are increasingly expected to act as persistent assistants: they must remember user preferences, past decisions, evolving facts, prior recommendations, and temporally grounded events across many sessions. This requirement has motivated a broad family of long-term memory systems that externalize interaction histories into retrievable stores~\citep{park2023generative,zhong2024memorybank,packer2023memgpt}. Recent systems further improve efficiency by compressing dialogues into facts, summaries, linked memory units, or staged memory stores~\citep{chhablani2025mem0,xu2025amem,fang2025lightmem,liu2026simplemem}. However, results on long-term conversational benchmarks such as LoCoMo and LongMemEval show that agent memory is not merely a problem of storing more text or retrieving semantically similar passages~\citep{maharana2024locomo,wu2024longmemeval}. Queries often depend on \emph{when} something happened, \emph{why} a user changed a plan, \emph{where} an event occurred, \emph{which} person or object was involved, or whether the user is asking about a previous assistant response.

\begin{figure}[t]
    \centering
    \includegraphics[width=\linewidth]{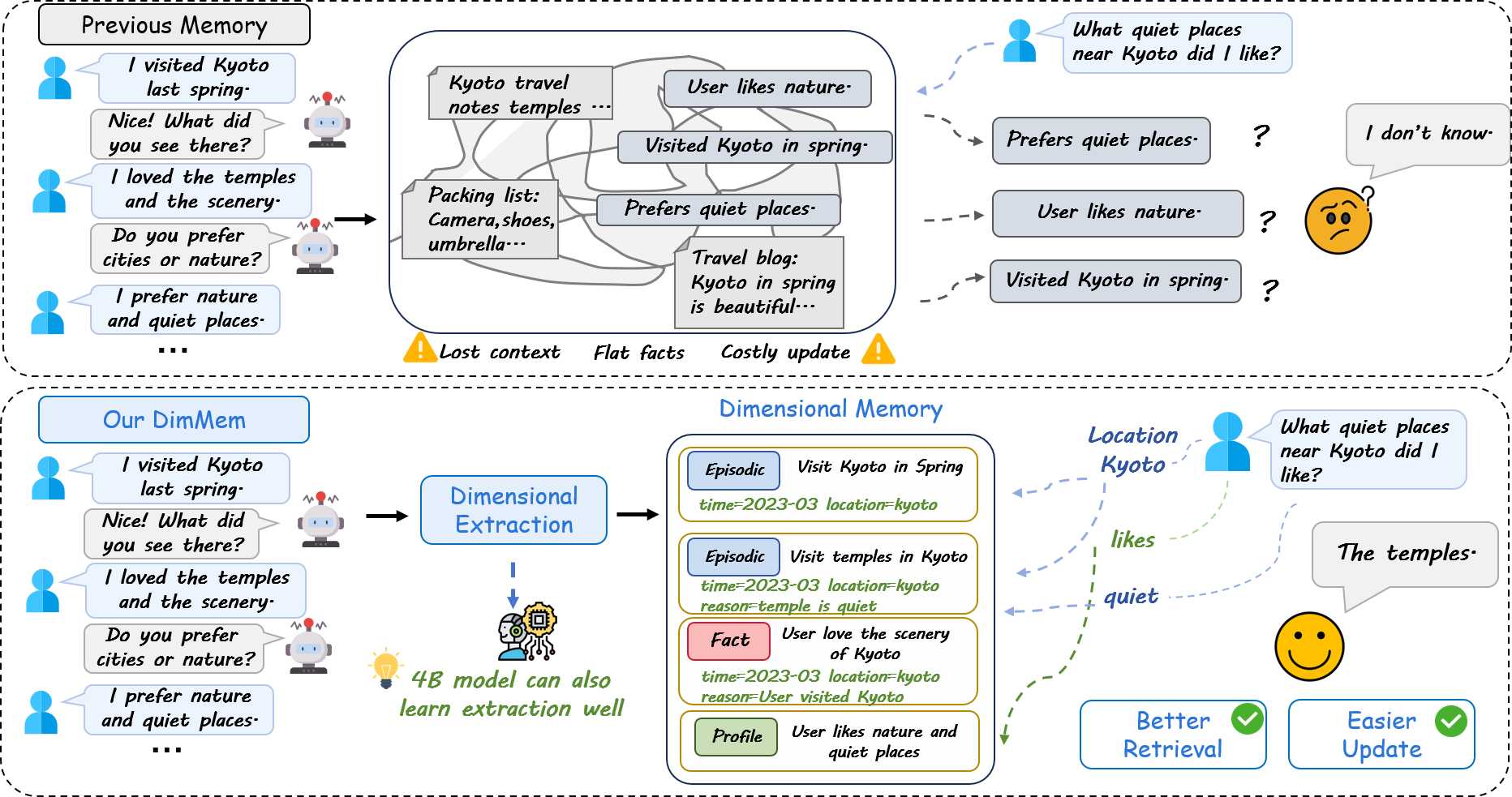}
    \caption{
    Conceptual motivation of DimMem. Prior memory systems often store conversations as raw, flat, or entangled memory forms, making retrieval and update depend on incomplete or costly context. DimMem instead exposes key dimensions within each memory unit, enabling more direct retrieval, easier update, and compact extractor learning.
    }
    \label{fig:motivation}
    \vspace{-1.4em}
\end{figure}

Existing memory systems therefore face a fidelity--efficiency dilemma. Raw dialogue or large text chunks preserve context, but they are expensive to store, retrieve, update, and place into the model context. Flat extracted facts are efficient, but they often discard the temporal, causal, intentional, and contextual information needed for precise recall. More structured designs, such as graph-based or hierarchical memory systems, can recover relational structure, but they introduce additional construction and maintenance cost~\citep{wu2026gam,zhang2026himem,xu2026structmem,lv2026allmem}. A parallel line of work learns memory operations through parameterized policies, using supervised fine-tuning or reinforcement learning to decide when to create, read, update, or delete memories~\citep{huo2026atommem,yu2026agenticmemory,yan2025memoryr1}. These approaches improve the control policy of memory management, but they still depend on the quality of the memory representation being controlled. If the stored unit does not expose time, reason, purpose, or memory type, downstream retrieval and update policies must infer these missing dimensions from an impoverished record.

Figure~\ref{fig:motivation} summarizes our central observation: long-term agent memory should be structured before it is retrieved, updated, or learned by a memory controller. Inspired by cognitive distinctions among episodic, semantic, and personal semantic memory~\citep{tulving1985memory,tulving2002episodic,renoult2012personal} and by event-oriented 5W1H information structure~\citep{hamborg2019giveme5w1h}, we propose \textbf{DimMem}, a lightweight dimensional memory framework for LLM agents. DimMem represents each memory as an atomic, self-contained unit with an explicit memory type (\texttt{fact}, \texttt{episodic}, or \texttt{profile}) and decoupled dimensional fields including \emph{time}, \emph{location}, \emph{reason}, \emph{purpose}, and typed \emph{keywords}. Rather than treating metadata as optional annotations, DimMem makes these fields the operational interface of the memory system.

This dimensional representation enables three system-level capabilities. First, DimMem performs dimension-aware retrieval by parsing a user query into the same schema as memory records and combining lexical matching, smantic matching, and dimension scoring. This tri-path retrieval prevents semantically related but temporally or causally mismatched memories from dominating the context. Second, DimMem supports dimension-aware update: memory type, keywords, temporal order, and causal/purpose fields provide cheap structural filters for identifying redundant or conflicting memories before expensive embedding or LLM-based consolidation. Third, DimMem introduces dynamic assistant-memory recall. Assistant replies are not blindly embedded into every memory record; instead, the query parser predicts whether assistant context is needed and retrieves the corresponding source replies only on demand.

DimMem also changes what is learned by a small model. Instead of training a model to choose memory CRUD actions, we train a compact extractor to produce high-quality dimensional memory records. This is important because memory extraction quality determines the substrate available to every later retrieval and update step. We fine-tune a 4B model on teacher-generated dimensional extraction examples and obtain a local structured extractor that learns the schema rather than merely imitating surface summaries. In our experiments, this fine-tuned 4B extractor reaches performance comparable to a much larger Qwen3-30B-A3B extractor, and in some settings even surpasses it, showing that dimensional memory extraction is a learnable capability rather than a privilege of proprietary or large-scale models.

Empirically, DimMem demonstrates that explicit dimensional structure is a strong inductive bias for agent memory. Across LoCoMo-10 and LongMemEval-S, DimMem consistently outperforms existing lightweight memory systems, with the largest gains appearing on dimension-dependent cases such as temporal reasoning, multi-session reasoning, knowledge updates, and assistant-dependent recall. Ablations further confirm that the dimensional schema, dimension retrieval route, and reason/purpose fields are not cosmetic metadata: they directly improve retrieval precision, memory update quality, and context efficiency.

Our contributions are summarized as follows:
\begin{itemize}[leftmargin=*,itemsep=2pt]
    \item \textbf{Dimensional memory representation.} We introduce DimMem, a lightweight long-term memory framework that represents each memory as an atomic, typed, and dimensionally structured record rather than a flat passage, summary, or isolated fact.
    \item \textbf{Dimension-aware memory operations.} We design retrieval, update, and assistant-recall mechanisms that align parsed query intents with explicit memory dimensions, improving temporal, causal, multi-session, and assistant-dependent reasoning while reducing retrieval and update cost.
    \item \textbf{Learnable structured extraction.} We show that high-quality dimensional memory extraction can be learned by a compact SFT model, and that extraction quality is crucial to end-to-end memory performance: a 4B extractor can approach or surpass larger general-purpose extractors when trained on the DimMem schema.
\end{itemize}

\section{Related Work}
\label{sec:related_work}

\subsection{Memory Systems for LLM Agents}
Long-term memory has long been studied through neural memory and retrieval-augmented models~\citep{weston2015memory,sukhbaatar2015endtoend,guu2020realm,khandelwal2020generalization,borgeaud2022improving}. For LLM agents, however, memory must support not only factual recall but also personalization, temporal continuity, preference tracking, and knowledge updates across interactions. Existing agent memory systems externalize dialogue history into retrievable stores, ranging from archival memory and personalized memory banks to fact-based stores, OS-style memory management, associative memories, staged updates, and query-planned retrieval~\citep{park2023generative,zhong2024memorybank,packer2023memgpt,chhablani2025mem0,kang2025memoryos,li2025memos,xu2025amem,fang2025lightmem,liu2026simplemem,hu2026xmemory}. Benchmarks such as LoCoMo and LongMemEval further show that long-term conversational memory requires temporal reasoning, multi-session tracking, knowledge updates, and assistant-side recall rather than simple semantic lookup~\citep{maharana2024locomo,wu2024longmemeval}. DimMem is positioned within this line of work, but improves the memory interface by making type, time, location, reason, purpose, and keywords explicit dimensions shared by extraction, retrieval, and update.

\subsection{Parameterized Memory Learning}
A growing line of work studies whether memory behavior should be learned rather than specified entirely by hand-written rules. Recent approaches use supervised fine-tuning and reinforcement learning to train memory managers that decide when to create, read, update, delete, or ignore memories~\citep{huo2026atommem,yu2026agenticmemory,yan2025memoryr1,zhang2026learningtoremember}. This trend reflects an important shift from static memory pipelines toward adaptive memory control, where the model learns operational decisions from data. Most of these methods parameterize the \emph{control policy}: they focus on which memory action to execute and when.
DimMem instead parameterizes a complementary component, the \emph{memory construction function}: it uses SFT to teach a compact extractor to produce high-quality dimensional memory records, improving the structured substrate on which later retrieval, update, or learned memory policies operate.

\subsection{Memory Storage Design and Management}
Memory systems also differ in how they store and manage accumulated experience. Lightweight designs store short facts, self-contained memory units, or topic-aware segments to reduce context and storage cost~\citep{fang2025lightmem,liu2026simplemem}. These representations are efficient, but their stored units often expose limited structure for temporal validity, causal background, task intent, or type-conditioned update behavior.
Richer designs organize memories with graphs, hierarchies, associative links, event structures, note abstractions, or topology-structured banks~\citep{xu2025amem,wu2026gam,zhang2026himem,xu2026structmem,lv2026allmem}. These approaches can strengthen relational reasoning and consolidation, but they also introduce graph construction, maintenance, and delayed-update overhead. DimMem takes a lighter-weight middle path: it avoids full graph or hierarchy construction, but embeds the structure needed for precise retrieval and update directly inside each atomic memory through explicit dimensions.

\section{Methodology}
\label{sec:methodology}

\begin{figure}[H]
    \centering
    \includegraphics[width=\textwidth]{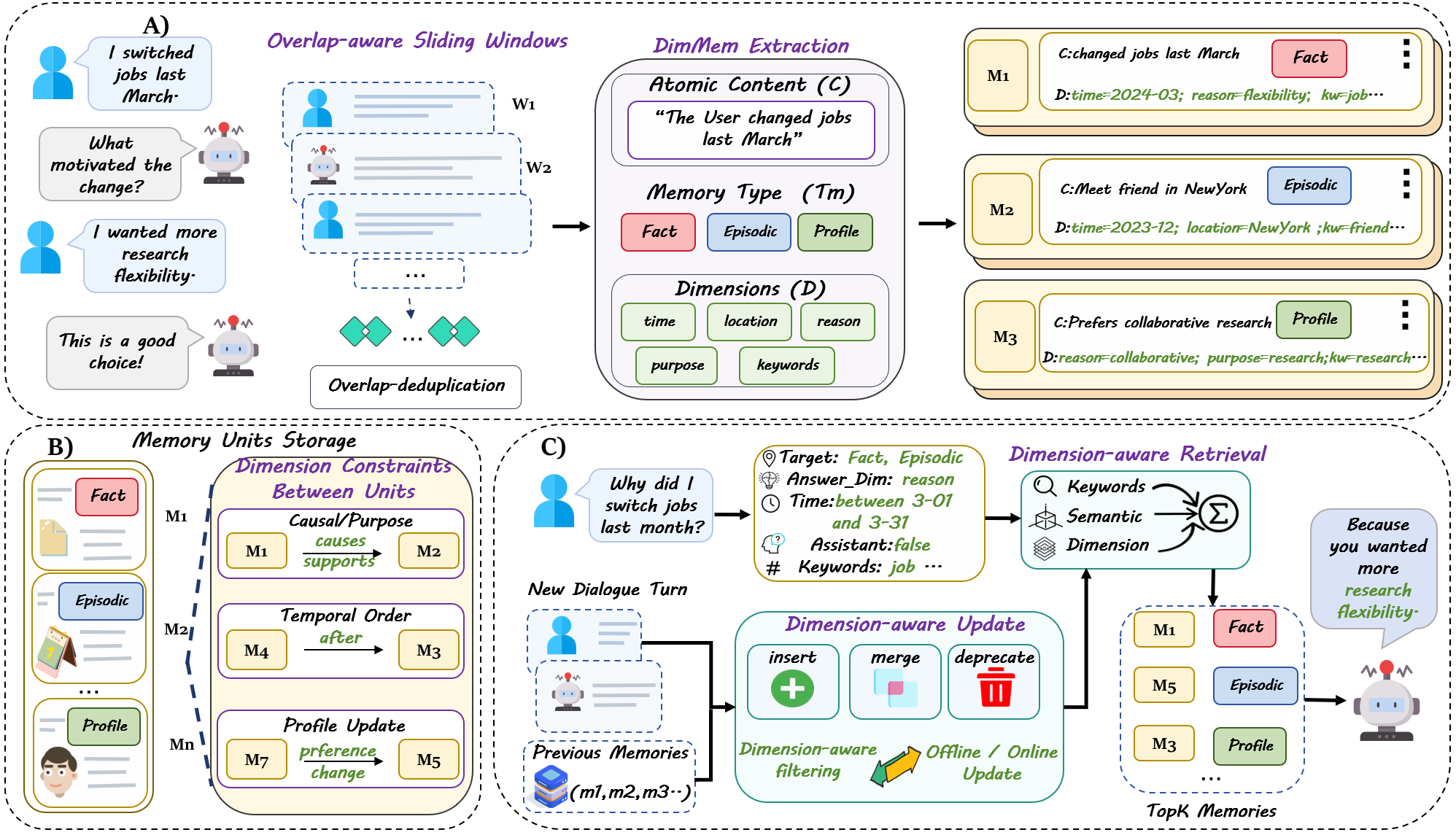}
\caption{
Overview of \textbf{DimMem}. 
(a) \textbf{Construction}: dialogue streams are segmented with overlap-aware windows and converted into atomic, typed, and dimensionally structured memory records. 
(b) \textbf{Storage}: explicit types and dimensions make cross-memory relations more discoverable, including temporal order, causal/purpose dependencies, and profile changes. 
(c) \textbf{Retrieval and update}: parsed queries use the same schema for tri-path retrieval, selective assistant-reply reconstruction, and dimension-aware insertion, merging, and deprecation.
}
    \label{fig:dimmem_framework}
    \vspace{-1.2em}
\end{figure}

DimMem follows a single principle: \emph{operational memory dimensions are exposed at construction time}, as shown in Figure~\ref{fig:dimmem_framework}. 
Instead of storing dialogues as raw passages, flat facts, or opaque embeddings, DimMem converts interactions into atomic records with explicit dimensions. 
User queries are parsed into the same schema, enabling retrieval, assistant-context reconstruction, and memory update to share a unified interface. 
Thus, what to store, how to retrieve, and how to maintain memory are coupled through the same dimensional representation.

\subsection{Dimensional Memory Records}
\label{sec:structure}

Let a dialogue stream be $\mathcal{S}=\{(u_t,a_t,\tau_t)\}_{t=1}^{T}$, where $u_t$, $a_t$, and $\tau_t$ denote the user utterance, assistant utterance, and timestamp at turn $t$. DimMem maps $\mathcal{S}$ into a memory bank $\mathcal{B}=\{m_i\}_{i=1}^{M}$, where each memory is represented as
\begin{equation}
    m_i = \langle id_i, C_i, T_i, \mathcal{D}_i, \rho_i \rangle ,
\end{equation}
where $id_i$ is a unique identifier, $C_i$ is a self-contained textual memory, $T_i$ is the memory type, $\mathcal{D}_i$ contains explicit dimensions, and $\rho_i$ points to the source window and turn. The pointer is not inserted into the QA context by default; it preserves access to grounded evidence, such as the original assistant reply, when a query explicitly requires it.

The type $T_i\in\{\texttt{fact},\texttt{episodic},\texttt{profile}\}$ distinguishes stable knowledge, event-specific experience, and long-term personal semantics~\citep{tulving1985memory,renoult2012personal}. The dimensional field is
\begin{equation}
    \mathcal{D}_i =
    \{D_{\mathrm{time}},D_{\mathrm{loc}},D_{\mathrm{reason}},
      D_{\mathrm{purpose}},D_{\mathrm{kw}}\}.
\end{equation}
Here $D_{\mathrm{time}}$ records when a memory is valid or occurred, $D_{\mathrm{loc}}$ records physical or contextual location, $D_{\mathrm{reason}}$ records causal background, $D_{\mathrm{purpose}}$ records communicative or task intent, and $D_{\mathrm{kw}}$ contains entity and topic anchors. Empty fields are allowed when unsupported by the dialogue. This convention avoids hallucinated metadata while still exposing grounded structure whenever the conversation provides it.

\subsection{Overlap-Aware Memory Extraction}
\label{sec:extraction}

Long dialogues are costly to process as a single prompt and may dilute memory-relevant signals. 
Inspired by prompt compression and lightweight memory-augmented generation~\citep{jiang-etal-2023-llmlingua,fang2025lightmem}, DimMem first compresses dialogue content before memory extraction. 
This reduces token consumption while filtering redundant or task-irrelevant information, thereby improving the signal-to-noise ratio of extraction. 
The dialogue $\mathcal{S}$ is then partitioned into overlapping windows:
\begin{equation}    
W_n = \{(u_t,a_t,\tau_t)\}_{t=s_n}^{e_n},    
\qquad    
s_{n+1}=e_n-O+1,
\end{equation}
where $O$ denotes the overlap size. 
The overlap preserves local discourse context across adjacent windows, but should not introduce duplicate memories. 
We therefore impose an extraction invariant: tokens in the overlap may be used to resolve coreference, normalize relative time, and identify implicit event boundaries, whereas newly extracted memories must be grounded in the non-overlapping suffix:
\begin{equation}    
\mathcal{M}_n =    
f_\theta\!\left(W_n;\operatorname{context}(W_n\cap W_{n-1})\right),    
\quad    
\operatorname{source}(m_i)\in W_n\setminus W_{n-1}.
\end{equation}

For each extracted memory, the extractor outputs a normalized record containing 
\texttt{content}, \texttt{memory\_type}, \texttt{time}, \texttt{location}, \texttt{reason}, \texttt{purpose}, and \texttt{keywords}. 
Each memory content $C_i$ must be self-contained: pronouns are resolved, relative temporal expressions are normalized using $\tau_t$, and unsupported dimensions are left empty. 
This yields a compact memory bank whose entries remain grounded through $\rho_i$ while remaining queryable without replaying the raw dialogue.

Following the trade-off analysis in \secref{app:segmentation_window_size}, we use window/overlap sizes of $15/3$ for LongMemEval and $25/5$ for LoCoMo to match their different turn densities.

\subsection{Intent Recognition and Dimension-Aligned Retrieval}
\label{sec:retrieval}

At inference time, DimMem maps a natural-language question into an operational memory intent before retrieval. This step identifies the requested answer dimension, the relevant memory type, the entity anchors, the dimensional constraints, and whether the user is asking about previous assistant behavior. Formally, given a query $q$, DimMem infers
\begin{equation}
    \hat{q} =
    \langle A_q, T_q, \tilde{\mathcal{D}}_q, E_q, C_q, I_{\mathrm{ast}}\rangle ,
\end{equation}
where $A_q$ denotes the answer dimension to recover, $T_q$ is the target memory type when identifiable, $\tilde{\mathcal{D}}_q$ contains constraints over the same dimensions used by memory records, $E_q$ contains grounded entity anchors, $C_q$ is a semantic query anchor, and $I_{\mathrm{ast}}\in\{0,1\}$ marks assistant-dependent intent. This representation separates \emph{what to answer} from \emph{which memories should be retrieved}. For example, a temporal constraint such as ``after March'' restricts retrieval, whereas a temporal answer request such as ``when did this happen'' sets $A_q=D_{\mathrm{time}}$ without filtering out memories that contain the answer.

\paragraph{Tri-path retrieval.}
DimMem uses three complementary retrieval views. Lexical matching is reliable for exact entities and rare phrases, dense retrieval recovers semantic paraphrases, and dimension matching enforces explicit constraints over type, time, location, reason, or purpose. Prior retrieval-augmented systems commonly combine multiple recall signals to compensate for the blind spots of individual retrievers~\citep{robertson2009bm25,reimers2019sentencebert,karpukhin2020dpr,lewis2020rag}. DimMem adopts this multi-route strategy, but makes the third route a \emph{dimension route} over the fields created by the extractor:
\begin{equation}
    \mathcal{R}(q)
    =
    \operatorname{Dedup}\!\left(
    \mathcal{R}_{\mathrm{bm25}}(q)
    \cup
    \mathcal{R}_{\mathrm{dim}}(q)
    \cup
    \mathcal{R}_{\mathrm{dense}}(q)
    \right).
\end{equation}
The lexical route $\mathcal{R}_{\mathrm{bm25}}$ matches query anchors against memory content and textual dimensions. The dense route $\mathcal{R}_{\mathrm{dense}}$ embeds query and memory text to recover paraphrases. The dimension route $\mathcal{R}_{\mathrm{dim}}$ scores explicit agreement between query constraints and memory dimensions:
\begin{equation}
    s_{\mathrm{dim}}(q,m_i)
    =
    \sum_{r\in\mathcal{A}(q)}
    \bar{\lambda}_{r}\,s_r(q,m_i),
    \qquad
    \bar{\lambda}_{r}
    =
    \frac{\lambda_r}{\sum_{r'\in\mathcal{A}(q)}\lambda_{r'}} .
\end{equation}
$\mathcal{A}(q)$ is the set of active constraints in the query, and weights are normalized over active dimensions so that absent fields do not dilute the score. The final list preserves evidence from all three routes while removing duplicate memory contents. This design makes dimensional structure operational: it does not replace lexical or semantic retrieval, but constrains them when the query specifies when, where, why, or what type of memory is needed.

\paragraph{Assistant-dependent intent.}
The intent variable $I_{\mathrm{ast}}$ controls whether the retrieval context should include previous assistant replies. Some questions ask what the assistant recommended, explained, or decided, while most questions can be answered from user-side memories alone. Storing all assistant replies in every memory record would inflate the memory bank and QA context, so DimMem treats assistant content as recoverable evidence. If $I_{\mathrm{ast}}=1$, each retrieved memory $m_i$ is mapped back through $\rho_i$, and the corresponding assistant reply is attached:
\begin{equation}
    \mathcal{C}_{\mathrm{aug}} =
    \{m_i \oplus \Phi_{\mathrm{ast}}(\rho_i)
    \mid m_i\in \mathcal{R}(q), I_{\mathrm{ast}}=1\},
\end{equation}
where $\Phi_{\mathrm{ast}}$ retrieves the original assistant utterance. If $I_{\mathrm{ast}}=0$, only structured memory records are used. Dynamic assistant recall is therefore part of intent handling: assistant context remains available when the intent requires it, but does not become a constant cost for all queries.

\subsection{Dimension-Aware Memory Update}
\label{sec:update}
As the memory bank grows, new records must be inserted, merged, used to supersede outdated records, or kept separate. Purely embedding-based updates may miss conflicts with different wording or over-merge semantically similar but temporally distinct memories. DimMem therefore uses dimensions as a cheap candidate-generation layer: records are first grouped by memory type and filtered by typed keyword overlap, then verified by embedding similarity before LLM consolidation. Thus, expensive LLM decisions are invoked only for structurally plausible candidate pairs.

For each candidate pair $(m_{\mathrm{new}},m_{\mathrm{old}})$, an LLM-based consolidator selects one action:
\begin{equation}
a\in\{\textsc{Merge},\textsc{Supersede},\textsc{KeepBoth}\}.
\end{equation}
The decision prompt exposes both content and dimensions, with type-conditioned semantics: for \texttt{fact} records, newer contradictory facts may \textsc{Supersede} older ones, while complementary facts may \textsc{Merge}; for \texttt{episodic} records, repeated descriptions of the same event may \textsc{Merge}, whereas distinct events, especially those occurring at different times, should \textsc{KeepBoth}; for \texttt{profile} records, overlapping preferences, habits, or style descriptions may \textsc{Merge}, while unrelated profile aspects remain separate. This update policy keeps the memory bank compact without erasing distinct episodes, and makes consolidation auditable by grounding decisions in explicit dimensions rather than hidden embedding similarity alone.

\subsection{Learning the Structured Extractor}
\label{sec:sft}

DimMem's schema is useful only if it can be populated reliably. We therefore train a compact extractor to map dialogue windows into structured memory records, rather than relying on a large proprietary model at inference time. To obtain supervision without using benchmark test annotations, we construct synthetic training data with GPT-5.4 as the teacher. Specifically, we generate 5K LoCoMo-style examples and 5K LongMemEval-style examples. The LongMemEval-style examples are generated from the UltraChat training split~\citep{ding2023enhancing}, by sampling dialogue windows and annotating them according to the DimMem schema. We filter malformed, empty, or schema-inconsistent outputs before fine-tuning.

We fine-tune \texttt{Qwen3-4B}~\citep{yang2025qwen3} with LoRA~\citep{hu2022lora}. The system prompt defines the memory taxonomy, field meanings, time normalization, coreference resolution, and the rule that unsupported dimensions remain empty. Each training sample follows the chat-completion format:
\begin{equation}    
x_i =    
[s_{\mathrm{schema}}, W_i, \mathcal{M}^{\mathrm{teacher}}_i],
\end{equation}
where $W_i$ is the dialogue window and $\mathcal{M}^{\mathrm{teacher}}_i$ is the teacher-generated memory JSON. After fine-tuning, the LoRA adapters are merged into the base model and served with vLLM~\citep{kwon2023efficient}. The trained extractor is then used as a drop-in component in the same DimMem pipeline, allowing us to test whether dimensional memory construction can be learned by a compact model rather than only elicited through large-model prompting.

\section{Experiments}
\label{sec:experiments}

\subsection{Experimental Setup}
\label{sec:exp_setup}

\textbf{Benchmarks.} We evaluate DimMem on two long-term conversational memory benchmarks: \textbf{LoCoMo-10}~\citep{maharana2024locomo}, which emphasizes temporal, multi-hop, open-domain, and single-hop reasoning over very long dialogues, and \textbf{LongMemEval-S}~\citep{wu2024longmemeval}, which stresses knowledge updates, multi-session tracking, temporal reasoning, user-preference recall, and assistant-side recall.

\textbf{Implementation Details.} We distinguish the \emph{memory extractor} from the \emph{QA model} to test whether DimMem's schema can be learned by a compact extractor without changing downstream QA. Unless otherwise specified, dense retrieval uses \texttt{MiniLM} sentence embeddings~\citep{wang2020minilm,reimers2019sentencebert}, and evaluation uses \texttt{GPT-4.1-mini} as the LLM judge~\citep{zheng2023judging}. We compare against raw-context, retrieval-based, and memory-system baselines, including MemoryOS~\citep{kang2025memoryos}, MemOS~\citep{li2025memos}, A-Mem~\citep{xu2025amem}, LightMem~\citep{fang2025lightmem}, and SimpleMem~\citep{liu2026simplemem}. \textbf{DimMem-4B} uses a fine-tuned \texttt{Qwen3-4B} extractor with \texttt{GPT-4.1-mini} for QA, while \textbf{DimMem-4B$\dagger$} uses fine-tuned \texttt{Qwen3-4B} for extraction and \texttt{Qwen3-4B} for QA, representing a fully compact local setting. All experiments are conducted on a single NVIDIA A800 GPU (80GB).

\subsection{Results and Analysis}
\label{sec:main_results}

\begin{table*}[t]
\centering
\caption{QA accuracy on the LoCoMo-10 benchmark. Best and second-best results within each model block are marked in \best{bold} and \second{underline}, respectively. For Tokens/Query, lower is better.}
\label{tab:locomo10}
\footnotesize
\setlength{\tabcolsep}{4.6pt}
\renewcommand{\arraystretch}{1.08}

\begin{NiceTabular*}{\textwidth}{@{\extracolsep{\fill}}lcccccc@{}}
\CodeBefore
\rowcolor{GPTHead}{2}
\rowcolor{GPTBlock}{3}
\rowcolor{GPTBlock}{4}
\rowcolor{GPTBlock}{5}
\rowcolor{GPTBlock}{6}
\rowcolor{GPTBlock}{7}
\rowcolor{GPTBlock}{8}
\rowcolor{GPTBlock}{9}
\rowcolor{GPTBlock}{10}
\rowcolor{GPTBlock}{11}
\rowcolor{QwenHead}{12}
\rowcolor{QwenBlock}{13}
\rowcolor{QwenBlock}{14}
\rowcolor{QwenBlock}{15}
\rowcolor{QwenBlock}{16}
\rowcolor{QwenBlock}{17}
\rowcolor{QwenBlock}{18}
\rowcolor{QwenBlock}{19}
\rowcolor{QwenBlock}{20}
\rowcolor{QwenBlock}{21}
\rowcolor{QwenBlock}{22}
\Body
\toprule
\textbf{Method} 
& \textbf{MultiHop} 
& \textbf{Temporal} 
& \textbf{OpenDomain} 
& \textbf{SingleHop} 
& \textbf{Overall} 
& \textbf{Tokens/Query} \\
\midrule
\multicolumn{7}{c}{\modelwithlogo{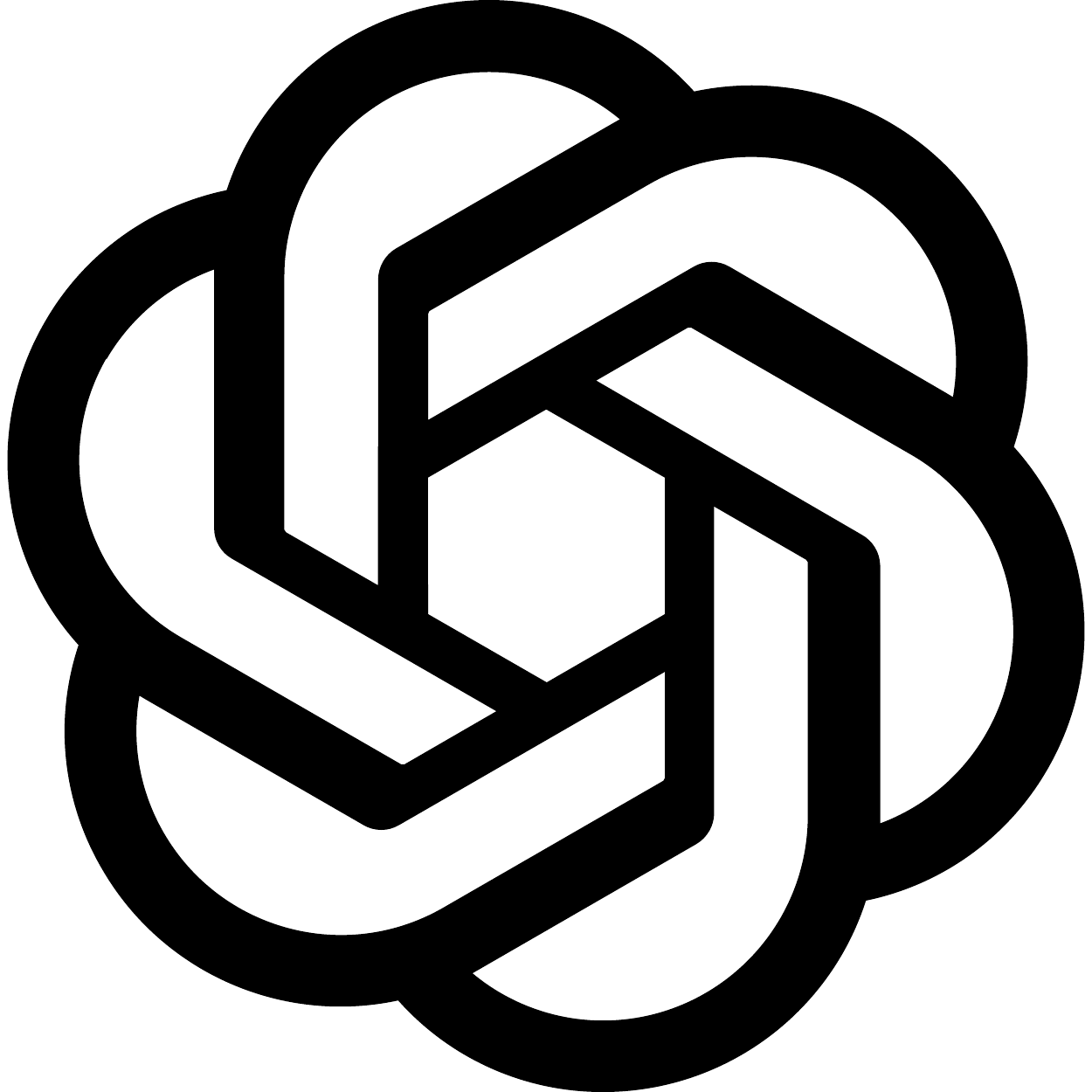}{GPT-4.1-mini}} \\
\midrule
FullText  & 59.22 & 68.22 & 50.00 & 72.41 & 67.72 & - \\
NaiveRAG  & 61.35 & 37.07 & 41.67 & 74.67 & 62.34 & 10,234 \\
MemoryOS  & 62.77 & 39.25 & 43.75 & 76.69 & 64.28 & 8,798 \\
MemOS     & 70.57 & 40.19 & 48.96 & 82.28 & 69.29 & 7,457 \\
A-Mem     & 63.12 & 40.81 & 50.00 & 78.00 & 65.78 & 9,743 \\
LightMem  & 62.77 & 64.17 & 54.17 & 81.57 & 72.79 & \second{5,063} \\
SimpleMem & 68.44 & 63.86 & \best{59.38} & \best{86.44} & 76.75 & 6,527 \\
\cdashline{1-7}
\textbf{DimMem (Ours)} 
& \best{81.91} & \best{81.62} & 55.21 & 82.50 & \second{80.51} & \best{3,859} \\
\textbf{DA-Update} 
& \second{80.85} & \second{81.00} & \second{55.21} & \second{84.78} & \best{81.43} &  - \\

\midrule
\multicolumn{7}{c}{\modelwithlogo{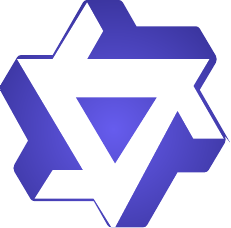}{Qwen3-30B-A3B-Instruct-2507}} \\
\midrule
FullText  & 56.74 & 46.73 & \second{57.29} & \best{87.75} & 71.62 & - \\
NaiveRAG  & 57.45 & 40.19 & 55.21 & 77.29 & 64.55 & 11,243 \\
MemoryOS  & 56.03 & 42.99 & 47.92 & 69.32 & 60.06 & 8,537 \\
MemOS     & 59.93 & 39.56 & 53.13 & 77.65 & 64.94 & 7,537 \\
A-Mem     & 53.19 & 41.12 & 55.21 & 66.11 & 57.86 & 10,347 \\
LightMem  & 61.70 & 60.75 & 51.04 & 79.31 & 70.45 & \second{6,721} \\
SimpleMem & 64.54 & 62.31 & 56.25 & 80.26 & 72.14 & 6,993 \\
\cdashline{1-7}
\textbf{DimMem (Ours)} 
& \second{75.53} & 69.16 & \best{58.33} & 81.09 & \second{76.17} & \best{5,846} \\
\textbf{DimMem-4B} 
& \best{76.95} & \best{81.31} & 57.29 & \second{82.88} & \best{79.87} & - \\
\textbf{DimMem-4B$\dagger$} 
& 72.70 & \second{75.70} & 46.88 & 76.10 & 73.57 & - \\
\bottomrule
\end{NiceTabular*}
\end{table*}

\begin{table*}[t]
\centering
\caption{QA accuracy on the LongMemEval-S benchmark. Best and second-best results within each model block are marked in \best{bold} and \second{underline}, respectively.}
\label{tab:longmemeval}
\footnotesize
\setlength{\tabcolsep}{4.6pt}
\renewcommand{\arraystretch}{1.08}
\begin{NiceTabular*}{\textwidth}{@{\extracolsep{\fill}}lccccccc@{}}
\CodeBefore
\rowcolor{GPTHead}{2}
\rowcolor{GPTBlock}{3}
\rowcolor{GPTBlock}{4}
\rowcolor{GPTBlock}{5}
\rowcolor{GPTBlock}{6}
\rowcolor{GPTBlock}{7}
\rowcolor{GPTBlock}{8}
\rowcolor{GPTBlock}{9}
\rowcolor{GPTBlock}{10}
\rowcolor{GPTBlock}{11}
\rowcolor{QwenHead}{12}
\rowcolor{QwenBlock}{13}
\rowcolor{QwenBlock}{14}
\rowcolor{QwenBlock}{15}
\rowcolor{QwenBlock}{16}
\rowcolor{QwenBlock}{17}
\rowcolor{QwenBlock}{18}
\rowcolor{QwenBlock}{19}
\rowcolor{QwenBlock}{20}
\rowcolor{QwenBlock}{21}
\rowcolor{QwenBlock}{22}
\Body
\toprule
\textbf{Method} 
& \textbf{TR} 
& \textbf{MS} 
& \textbf{KU} 
& \textbf{SSU} 
& \textbf{SSA} 
& \textbf{SSP} 
& \textbf{Overall} \\
\midrule

\multicolumn{8}{c}{\modelwithlogo{openai}{GPT-4.1-mini}} \\
\midrule
FullText  & 31.58 & 46.62 & 43.59 & 80.00 & 41.07 & 56.67 & 46.80 \\
NaiveRAG  & 57.14 & 54.14 & 74.36 & 77.14 & 60.71 & 53.33 & 62.00 \\
MemoryOS  & 50.38 & 48.12 & 71.79 & 85.71 & 51.79 & 66.67 & 59.20 \\
MemOS     & 63.91 & 54.14 & 70.51 & 90.00 & 64.29 & 50.00 & 65.20 \\
A-Mem     & 58.65 & 52.63 & 69.23 & 88.57 & 57.14 & 46.67 & 62.00 \\
SimpleMem & \second{81.95} & 57.89 & 74.36 & 82.86 & 71.43 & \second{76.67} & 73.00 \\
LightMem  & \best{84.21} & 51.13 & \second{88.46} & 90.00 & 23.21 & \best{76.67} & 69.60 \\
\cdashline{1-8}
\textbf{DimMem (Ours)} 
& 74.44 & 64.66 & 84.62 & \second{98.57} & \best{76.79} & 63.33 & \second{76.40} \\
\textbf{DA-Update} 
& 77.44 & \best{69.17} & \best{88.46} & \best{98.57} & \second{73.21} & 56.67 & \best{78.20} \\

\midrule
\multicolumn{8}{c}{\modelwithlogo{qwen}{Qwen3-30B-A3B-Instruct-2507}} \\
\midrule
FullText  & 63.91 & 42.11 & 65.38 & 71.43 & 32.14 & 46.67 & 54.80 \\
NaiveRAG  & 65.41 & 48.87 & 71.79 & 81.43 & 30.36 & 53.33 & 59.60 \\
MemoryOS  & 57.14 & 40.60 & 64.10 & 61.43 & 28.57 & 50.00 & 50.80 \\
MemOS     & 72.93 & 51.13 & 73.08 & 80.00 & 26.79 & 53.33 & 61.80 \\
A-Mem     & 66.92 & 44.36 & 71.79 & 74.29 & 35.71 & 63.33 & 59.00 \\
SimpleMem & 74.44 & 52.63 & 75.64 & \second{87.14} & \second{64.29} & \second{66.67} & \second{69.00} \\
LightMem  & \second{75.94} & \best{57.89} & \best{89.74} & 85.71 & 21.43 & \best{73.33} & 68.40 \\
\cdashline{1-8}
\textbf{DimMem (Ours)} 
& \best{76.69} & \second{54.89} & \second{79.49} & \best{94.29} & \best{73.21} & 56.67 & \best{72.20} \\
\textbf{DimMem-4B} 
& 73.68 & \second{67.42} & 83.33 & 90.00 & 46.43 & 60.00 & 71.80 \\
\textbf{DimMem-4B$\dagger$} 
& 76.69 & 55.64 & 83.33 & 84.29 & 42.86 & 56.67 & 68.20 \\
\bottomrule
\end{NiceTabular*}

\end{table*}

Tables~\ref{tab:locomo10} and~\ref{tab:longmemeval} present the main results. We focus on three questions: whether dimensional memories improve long-term reasoning, whether the update and assistant-recall mechanisms address benchmark-specific failure modes, and whether a compact model can learn to construct useful dimensional memories. KU=Knowledge-Update, MS=Multi-Session, SSA=Single-Session-Assistant, SSP=Single-Session-Preference, SSU=Single-Session-User, TR=Temporal-Reasoning.

\textbf{Dimensional memories improve long-term reasoning.} DimMem consistently outperforms memory baselines across both benchmarks. The gains are most visible on LoCoMo-10, where questions often require combining entities, events, and temporal context across long conversations. The result supports our central hypothesis: storing compact memories is not enough; the stored unit must expose the dimensions needed by later retrieval and reasoning.

\textbf{Dimension-aware operations target the hard cases.} DA-Update improves the cases where memories evolve or conflict over time, especially multi-session and knowledge-update questions. Dynamic assistant recall addresses a different failure mode: some queries ask about what the assistant previously recommended or explained. Instead of storing all assistant responses in every memory, DimMem retrieves assistant context only when the query requires it, which improves assistant-dependent recall without increasing the default memory context.

\textbf{Fine-tuning makes compact extraction viable.} DimMem-4B shows that the dimensional extraction skill can be learned by a small model. With the fine-tuned \texttt{Qwen3-4B} extractor and \texttt{GPT-4.1-mini} QA, DimMem-4B reaches performance comparable to, and in LoCoMo-10 even higher than, DimMem using the larger \texttt{Qwen3-30B-A3B} extractor. DimMem-4B$\dagger$ is lower because both extraction and QA are handled by \texttt{Qwen3-4B}, but it remains a useful fully compact configuration. This separation shows that much of the gain comes from learning to construct high-quality dimensional memories, while the remaining gap is largely due to the QA model.

The main results show that DimMem improves long-term conversational memory. This section isolates the design choices behind these gains: whether dimensional records improve the stored representation, whether the dimension route adds recall beyond lexical and dense retrieval, and whether dimensions reduce the cost of assistant recall and memory update. We further provide case studies in    \secref{app:case_study}. Additional segmentation and update analysis is provided in  \secref{app:analysis_update} and   \secref{app:segmentation_window_size}.

\subsection{Ablation Study}
\label{sec:ablation}

We evaluate three controlled variants using \texttt{GPT-4.1-mini} as the backbone model.  \textbf{Content-Only Memory} collapses each record into flat text, isolating the value of dimensional representation. \textbf{Lexical+Dense Retrieval} keeps dimensional records but removes dimension-based scoring, isolating the value of query--memory dimension alignment. \textbf{No Causal/Intent Dimensions} removes reason and purpose while preserving type, time, location, and keywords, isolating the contribution of causal and intentional fields.

\begin{table}[H]
\centering
\caption{Ablation results on LongMemEval-S (\%). Absolute changes are relative to DimMem (Full).}
\label{tab:ablation_longmemeval}
\small
\resizebox{\linewidth}{!}{
\begin{tabular}{lccccccc}
\toprule
\textbf{Config} & \textbf{KU} & \textbf{MS} & \textbf{SSA} & \textbf{SSP} & \textbf{SSU} & \textbf{TR} & \textbf{Overall} \\
\midrule
\rowcolor{gray!15}
DimMem (Full)
& 88.46
& 69.17
& 73.21
& 56.67
& 98.57
& 77.44
& \textbf{78.20} \\

Content-Only Memory
& 85.90\dropc{2.56}
& 61.65\dropc{7.52}
& 17.86\dropc{55.35}
& 63.33\gain{6.66}
& 98.57\nochange
& 75.94\dropc{1.50}
& 69.60\dropc{8.60} \\

Lexical+Dense Retrieval
& 87.18\dropc{1.28}
& 62.41\dropc{6.76}
& 76.79\gain{3.58}
& 60.00\gain{3.33}
& 97.14\dropc{1.43}
& 72.18\dropc{5.26}
& 75.20\dropc{3.00} \\

No Causal/Intent Dims.
& 88.46\nochange
& 69.92\gain{0.75}
& 76.79\gain{3.58}
& 53.33\dropc{3.34}
& 97.14\dropc{1.43}
& 72.18\dropc{5.26}
& 77.00\dropc{1.20} \\
\bottomrule
\end{tabular}
}
\end{table}

\begin{table}[H]
\centering
\caption{Ablation results on LoCoMo-10 (\%). Absolute changes are relative to DimMem (Full).}
\label{tab:ablation_locomo}
\small
\resizebox{\linewidth}{!}{
\begin{tabular}{lccccc}
\toprule
\textbf{Config} & \textbf{Multi-hop} & \textbf{Temporal} & \textbf{Open-domain} & \textbf{Single-hop} & \textbf{Overall} \\
\midrule
\rowcolor{gray!15}
DimMem (Full)
& 80.85
& 81.00
& 55.21
& 84.78
& \textbf{81.43} \\

Content-Only Memory
& 62.77\dropc{18.08}
& 64.17\dropc{16.83}
& 52.08\dropc{3.13}
& 78.60\dropc{6.18}
& 71.04\dropc{10.39} \\

Lexical+Dense Retrieval
& 81.02\gain{0.17}
& 77.00\dropc{4.00}
& 53.26\dropc{1.95}
& 83.47\dropc{1.31}
& 79.87\dropc{1.56} \\

No Causal/Intent Dims.
& 80.85\nochange
& 79.75\dropc{1.25}
& 55.21\nochange
& 83.69\dropc{1.09}
& 80.58\dropc{0.85} \\
\bottomrule
\end{tabular}
}
\end{table}

Tables~\ref{tab:ablation_longmemeval} and~\ref{tab:ablation_locomo} show that the representation is the dominant factor: Content-Only Memory causes the largest overall drop on both benchmarks, especially on LoCoMo-10 multi-hop and temporal questions and on LongMemEval-S assistant-dependent questions. Lexical+Dense Retrieval produces smaller but consistent losses, showing that BM25 and dense retrieval recover many candidates but cannot directly enforce time, type, and location constraints. Removing causal and intentional fields has the smallest overall effect, but still hurts temporal reasoning, suggesting that reason and purpose are most useful when lexically similar memories differ in cause or intent.

\subsection{Dynamic Assistant Recall: Selective Context Reconstruction}
\label{sec:ablation_dynamic_recall}

Some questions ask about previous assistant replies rather than user-stated facts. Storing all assistant replies in every memory record would inflate all queries, so DimMem predicts whether assistant context is needed and reconstructs it through source pointers only on demand.

\begin{figure}[H]
    \centering
    \includegraphics[width=\linewidth]{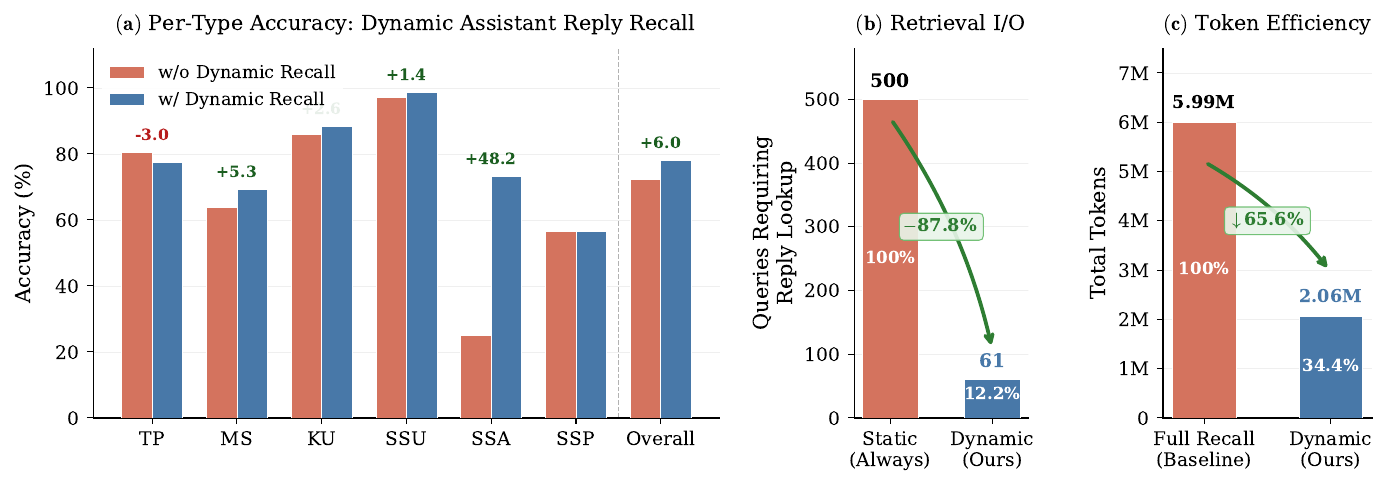}
    \caption{Dynamic assistant recall on LongMemEval-S. The mechanism improves assistant-dependent questions while triggering reply lookup for only a small fraction of all queries.}
    \label{fig:dynamic_recall}
\end{figure}

Figure~\ref{fig:dynamic_recall} shows that selective recall captures most of the benefit of assistant context at a much lower cost. It improves overall accuracy by 6.0 points, with the largest gain on Single-Session-Assistant questions (+48.2), while triggering assistant-reply lookup for only 12.2\% of queries. This reduces token consumption to 34.4\% of full recall (2.06M vs.\ 5.99M tokens). We further analyze the detection quality of the recall trigger in  \secref{app:dynamic_recall_detection}. The detector achieves 90.2\% precision, 98.2\% recall, and 94.0\% F1, missing only one assistant-dependent query, which explains why selective recall remains robust while avoiding unnecessary assistant-context reconstruction.

\subsection{Update Efficiency: Dimensions Reduce Candidate Search}
\label{app:analysis_update}

Memory update becomes expensive when every new record is compared against the full bank. We compare DimMem with a LightMem-style embedding baseline. DimMem first filters candidates by \texttt{memory\_type} and keyword overlap, then applies embedding verification and LLM consolidation only to the reduced set.

\begin{table}[H]
\centering
\caption{Memory update efficiency. DimMem uses dimensional filters before embedding and LLM consolidation.}
\label{tab:update_efficiency}
\small
\begin{tabular}{llcccc}
\toprule
\textbf{Dataset} & \textbf{Method} & \textbf{Embed. Count} & \textbf{Embed. Saved} & \textbf{Merges} & \textbf{Compression} \\
\midrule
\multirow{2}{*}{LongMemEval} & LightMem & 110,601 (100\%) & --- & 5,897 & 5.33\% \\
& \cellcolor{gray!15} DimMem & \cellcolor{gray!15} 27,107 (24.5\%) & \cellcolor{gray!15} \textbf{75.5\%} & \cellcolor{gray!15} \textbf{7,555} & \cellcolor{gray!15} \textbf{6.83\%} \\
\midrule
\multirow{2}{*}{LoCoMo} & LightMem & 2,475 (100\%) & --- & 241 & 9.74\% \\
& \cellcolor{gray!15} DimMem & \cellcolor{gray!15} 1,011 (40.8\%) & \cellcolor{gray!15} \textbf{59.2\%} & \cellcolor{gray!15} \textbf{247} & \cellcolor{gray!15} \textbf{9.98\%} \\
\bottomrule
\end{tabular}
\end{table}

Table~\ref{tab:update_efficiency} shows that dimensional filtering reduces embedding computation by 75.5\% on LongMemEval and 59.2\% on LoCoMo, while finding slightly more merges than the baseline. The extra merges come from pairs that share memory type and typed keywords but fall below the baseline's high embedding threshold. Thus, dimensions act as cheap filters that improve both efficiency and candidate quality before expensive LLM consolidation.

\subsection{Detection Quality of Dynamic Assistant Recall}
\label{app:dynamic_recall_detection}

To further examine whether dynamic assistant recall reliably identifies assistant-dependent questions, we report the detection confusion matrix based on GPT-4.1-mini in Table~\ref{tab:recall_confusion}. The ground truth is whether the query belongs to the assistant-dependent Single-Session-Assistant (SSA) type.

\begin{table}[H]
\centering
\caption{Detection quality for dynamic assistant recall on LongMemEval-S. Ground truth is whether the query belongs to the assistant-dependent SSA type.}
\label{tab:recall_confusion}
\small
\begin{tabular}{l|cc|c}
\toprule
& \textbf{Actual Need} & \textbf{Actual No Need} & \textbf{Total} \\
\midrule
\textbf{Predicted Need} & \textbf{55} & 6 & 61 \\
\textbf{Predicted No Need} & 1 & \textbf{438} & 439 \\
\midrule
\textbf{Total} & 56 & 444 & 500 \\
\bottomrule
\end{tabular}
\\[3pt]
\scriptsize{Precision = 90.2\%, Recall = 98.2\%, F1 = 94.0\%.}
\end{table}

Table~\ref{tab:recall_confusion} shows that the recall trigger is highly reliable. It misses only one assistant-dependent query and introduces only six false positives. The false positives mainly increase a small amount of irrelevant context, whereas the low false-negative rate ensures that nearly all assistant-dependent questions receive the required source-level reply reconstruction. This supports the design choice of storing source pointers rather than duplicating assistant replies inside every memory record.

\section{Conclusion and Limitations}
\label{sec:conclusion}

We presented \textbf{DimMem}, a lightweight dimensional memory framework for long-term LLM agents. DimMem stores memories as atomic, typed records with explicit dimensions, enabling extraction, query parsing, retrieval, assistant recall, and update to share one schema. Experiments on LoCoMo-10 and LongMemEval-S show improved accuracy and efficiency, while compact-extractor results suggest that dimensional memory construction can be learned by smaller models.
\textbf{Limitations.}DimMem is evaluated only on textual conversational memory, and should be further tested in deployed, multimodal, and privacy-sensitive agents. It also depends on accurate extraction and intent parsing, and its schema may require domain-specific extensions. Update consolidation still relies on LLM judgments.

\clearpage

\bibliographystyle{unsrt}
\bibliography{references}

\appendix

\begin{center}
    {\Large \bfseries APPENDIX}
\end{center}
\vspace{1em}

\section{Case Study}
\label{app:case_study}

We walk through a representative retrieval example end-to-end to illustrate three operational stages of DimMem: (1)~query parsing into explicit dimensions, (2)~dimension-aware scoring, and (3)~structured memory records presented to the reader model. This is followed by a second, shorter case showing how dimension-aware update resolves conflicting memory records.

\subsection{Case 1: End-to-End Dimension-Aware Retrieval}
\label{sec:case_retrieval}

\noindent\textbf{Query.}
\textit{``How many days did I spend on camping trips in the United States this year?''}\\
\textbf{Question date.} 2023/04/29 \quad
\textbf{Gold answer.} 8 days.

\subsubsection{Step 1: Dimensional Query Parsing}

The query analyzer maps the natural-language question into a structured retrieval specification. Below is the actual parsed output:

\begin{table}[h]
\centering
\caption{Parsed query dimensions produced by the query analyzer.}
\label{tab:case_parsed_query}
\small
\renewcommand{\arraystretch}{1.15}
\begin{tabular}{lp{9.5cm}}
\toprule
\textbf{Field} & \textbf{Parsed Value} \\
\midrule
\texttt{query\_anchor} & ``The number of days the user spent on camping trips in the United States in 2023'' \\
\texttt{target\_memory\_type} & [\texttt{episodic}] \\
\texttt{keywords} & [``camping trips'', ``United States''] \\
\texttt{time} & \texttt{between 2023-01-01 and 2023-04-29} \\
\texttt{location} & \texttt{United States} \\
\texttt{answer\_dim} & \texttt{content} \\
\texttt{need\_assistant\_context} & \texttt{false} \\
\bottomrule
\end{tabular}
\end{table}

\noindent The query analyzer makes several important decisions:
\begin{itemize}[leftmargin=*,itemsep=2pt]
    \item It converts the relative phrase ``this year'' to the constraint \texttt{between 2023-01-01 and 2023-04-29}, anchored to the question date.
    \item It identifies the memory type \texttt{episodic} (trip events) rather than \texttt{fact} or \texttt{profile}.
    \item It extracts ``United States'' as both a keyword and location constraint.
    \item It determines that the answer is in memory content (number of days) rather than in time or location fields.
\end{itemize}

\subsubsection{Step 2: Dimension-Aware Scoring}

The dimension route scores each memory record against the parsed constraints. Table~\ref{tab:case_scoring} shows the per-component breakdown for two candidate records. Only dimensions present in the query are activated.

\begin{table}[h]
\centering
\caption{Dimension scoring breakdown for two candidate memory records.}
\label{tab:case_scoring}
\small
\renewcommand{\arraystretch}{1.15}
\begin{tabular}{lcccc}
\toprule
\textbf{Component} & \textbf{Weight $\bar{\lambda}_r$} & \textbf{Record A} & \textbf{Record B} \\
\midrule
\texttt{memory\_type} (target: episodic)
    & 0.158 & 1.0 \cmark & 1.0 \cmark \\
\texttt{time\_constraint} (between 2023-01 and 2023-04-29)
    & 0.316 & 1.0 \cmark & 0.0 \xmark \\
\texttt{location} (United States)
    & 0.211 & 0.0 & 1.0 \cmark \\
\texttt{keyword\_phrase} (``camping trips'', ``United States'')
    & 0.158 & 0.0 & 0.0 \\
\texttt{keyword\_token\_overlap}
    & 0.158 & 0.50 & 0.25 \\
\midrule
\textbf{Final $s_{\mathrm{dim}}$} & --- & \textbf{0.553} & \textbf{0.408} \\
\bottomrule
\end{tabular}
\end{table}

\noindent\textbf{Record A} (rank 1 in dimension route): a 3-day camping trip to Big Sur in early April 2023. It scores high because its type is episodic and its source timestamp (2023-04-29) falls within the query's time window. The location field contains ``Big Sur'' but not the string ``United States'', so the location sub-score is 0---yet the time and type signals are strong enough to rank it above records without temporal grounding.

\noindent\textbf{Record B}: planning a trip to Rocky Mountains in Colorado (2023-04-29). It matches location (``Colorado'' is not the exact string ``United States'') but its source timestamp falls on the boundary. This illustrates how multi-component scoring selects records with the best aggregate dimensional fit rather than relying on any single dimension.

\subsubsection{Step 3: Retrieved Structured Memories}

Table~\ref{tab:case_records} shows the top-3 records from the dimension route, demonstrating how each record exposes its full dimensional schema.

\begin{table*}[t]
\centering
\caption{Top-3 records from the dimension route for the camping query. Each record carries the same dimensional schema used at write time, enabling structured matching at query time.}
\label{tab:case_records}
\small
\renewcommand{\arraystretch}{1.12}
\begin{tabular}{cp{7.8cm}p{5.8cm}}
\toprule
\textbf{\#} & \textbf{Content} & \textbf{Dimensions} \\
\midrule
1
& The user recently returned from a \textbf{3-day solo camping trip} to Big Sur in early April and found their hiking boots adequate but thinks they need waterproof boots for multi-day backpacking trips.
& \texttt{type}: episodic \newline
  \texttt{time}: early April 2023 \newline
  \texttt{location}: Big Sur \newline
  \texttt{reason}: need for better boots after trip \newline
  \texttt{purpose}: find waterproof boots for multi-day backpacking \newline
  \texttt{keywords}: [Big Sur, solo camping, hiking boots, waterproof boots, multi-day backpacking] \\
\midrule
2
& The user recently completed an amazing \textbf{5-day camping trip} to Yellowstone National Park in March 2023 and is still enthusiastic about the experience.
& \texttt{type}: episodic \newline
  \texttt{time}: 2023-03 \newline
  \texttt{location}: Yellowstone National Park \newline
  \texttt{reason}: --- \newline
  \texttt{purpose}: --- \newline
  \texttt{keywords}: [Yellowstone National Park, 5-day camping trip, outdoor experience] \\
\midrule
3
& The user was interested in visiting nearby national parks and requested recommendations for day trips near Moab, Utah.
& \texttt{type}: episodic \newline
  \texttt{time}: 2023-04-29 \newline
  \texttt{location}: Moab, Utah \newline
  \texttt{reason}: --- \newline
  \texttt{purpose}: --- \newline
  \texttt{keywords}: [national parks, day trips, Moab Utah, short trips] \\
\bottomrule
\end{tabular}
\end{table*}

\noindent\textbf{Contrast with BM25-only.}
Without the dimension route, BM25 ranks a Passover-related record at position 4 (matching ``number'' and common function words) and pushes the Yellowstone camping record lower. The dimension route's time constraint and type filter jointly ensure that both camping trip records appear in the top-5, enabling the reader model to sum the durations correctly ($3+5=8$ days).

\subsection{Case 2: Dimension-Aware Memory Update}
\label{sec:case_update}

\noindent\textbf{Scenario.}
During conversation, the user first expressed a general goal---reaching the Gold level on their Starbucks Rewards app---and later corrected the requirement as \textbf{125 stars} rather than 400. DimMem's update pipeline detects this conflict before retrieval.

\subsubsection{Multi-Level Candidate Detection}

\begin{table}[h]
\centering
\caption{Dimension-aware update: the multi-level filter identifies a candidate pair that purely embedding-based deduplication would miss.}
\label{tab:case_update}
\small
\renewcommand{\arraystretch}{1.15}
\begin{tabular}{p{2.0cm}p{11.0cm}}
\toprule
\textbf{Stage} & \textbf{Operation} \\
\midrule
\texttt{Level 1}: Type grouping
& Both records are \texttt{memory\_type}=\texttt{fact} $\to$ same-type candidate pair formed. \\
\midrule
\texttt{Level 2}: Keyword overlap
& Keywords intersection: \{``Starbucks'', ``Gold'', ``stars''\}; Jaccard $> 0.3$ $\to$ passed. \\
\midrule
\texttt{Level 3}: Embedding
& Cosine similarity $= 0.797$, above DimMem threshold $0.7$ $\to$ passed. \\
& \textit{(Note: LightMem uses threshold $0.85$; this pair would be missed.)} \\
\midrule
\texttt{Level 4}: LLM decision
& Same entity (Starbucks Gold level); values conflict (vague $\to$ 125). Action: \textbf{\textsc{Supersede}}. Old memory archived; corrected memory retained. \\
\bottomrule
\end{tabular}
\end{table}

\noindent\textbf{Before and after update.}
\begin{itemize}[leftmargin=*,itemsep=2pt]
    \item \textbf{Old:} ``The user is trying to reach the Gold level on their Starbucks Rewards app and wants to know how many stars are needed.''
    \item \textbf{New (retained):} ``The user corrected that \textbf{125 stars} are needed to reach Gold level on the Starbucks Rewards app, not 400.''
\end{itemize}

\subsubsection{Why Dimensions Matter for Update}

The multi-level pipeline uses three key observations:
\begin{enumerate}[leftmargin=*,itemsep=2pt]
    \item \textbf{Type grouping} eliminates the vast majority of pairwise comparisons: only records of the same memory type can conflict.
    \item \textbf{Keyword overlap} acts as a cheap entity-level filter: ``Starbucks'' + ``Gold'' + ``stars'' strongly suggests these two records are about the same entity.
    \item \textbf{A lower embedding threshold} ($0.7$ vs.\ the typical $0.85$) is safe because it is applied \emph{after} the keyword filter has already ensured topical relatedness.
\end{enumerate}

\noindent Without this multi-level design, either the pair goes undetected (threshold too high) or all memory pairs must be checked by the LLM consolidator (no keyword pre-filter). DimMem's cascaded approach achieves high recall on true conflicts while keeping LLM calls manageable.

\section{Hyperparameters}
                                                                                                    
\begin{table}[t]
\centering
\caption{Pipeline and model hyperparameters used in LongMemEval and LoCoMo.}
\label{tab:hyperparams}
\small
\setlength{\tabcolsep}{6pt}
\renewcommand{\arraystretch}{1.12}

\begin{tabularx}{\linewidth}{>{\raggedright\arraybackslash}Xcc}
\toprule
\textbf{Hyperparameter} 
& \textbf{LongMemEval} 
& \textbf{LoCoMo} \\
\midrule

\rowcolor{gray!12}
\multicolumn{3}{l}{\textbf{\textit{Segmentation}}} \\
Window size & 15 & 25 \\
Overlap & 3 & 5 \\

\addlinespace[2pt]
\rowcolor{gray!12}
\multicolumn{3}{l}{\textbf{\textit{Compression (LLMLingua-2)}}} \\
Compression rate & 0.8 & 0.8 \\
Compression model 
& \multicolumn{2}{c}{\makecell[c]{bert-base-multilingual-cased\\-meetingbank}} \\

\addlinespace[2pt]
\rowcolor{gray!12}
\multicolumn{3}{l}{\textbf{\textit{Memory Extraction}}} \\
LLM & \multicolumn{2}{c}{GPT-4.1-mini \/ Qwen3-30B-A3B-Instruct-2507 } \\
Temperature & 0.0 & 0.0 \\
Max output tokens & 16{,}384 & 16{,}384 \\

\addlinespace[2pt]
\rowcolor{gray!12}
\multicolumn{3}{l}{\textbf{\textit{Query Analysis}}} \\
Temperature & 0.0 & 0.0 \\
Max output tokens & 4{,}096 & 4{,}096 \\

\addlinespace[2pt]
\rowcolor{gray!12}
\multicolumn{3}{l}{\textbf{\textit{Retrieval}}} \\
Top-$k$ per route & 20 & 15 \\
Embedding model 
& \multicolumn{2}{c}{\makecell[c]{all-MiniLM-L6-v2 (384-dim)}} \\

\addlinespace[2pt]
\rowcolor{gray!12}
\multicolumn{3}{l}{\textbf{\textit{QA \& Judge}}} \\
LLM & \multicolumn{2}{c}{gpt-4.1-mini} \\
Temperature & 0.0 & 0.0 \\
Max output tokens (Judge) & 512 & 512 \\

\addlinespace[2pt]
\rowcolor{gray!12}
\multicolumn{3}{l}{\textbf{\textit{LoRA Fine-tuning (Qwen3-4B)}}} \\
Precision & \multicolumn{2}{c}{bfloat16} \\
Epochs & \multicolumn{2}{c}{2} \\
Per-device batch size & \multicolumn{2}{c}{1} \\
Gradient accumulation steps & \multicolumn{2}{c}{2} \\
Effective batch size & \multicolumn{2}{c}{2} \\
Learning rate & \multicolumn{2}{c}{$5 \times 10^{-5}$} \\
Max sequence length & \multicolumn{2}{c}{12{,}288} \\
LoRA rank ($r$) & \multicolumn{2}{c}{16} \\
LoRA alpha ($\alpha$) & \multicolumn{2}{c}{32} \\
\bottomrule
\end{tabularx}
\end{table}

\section{Additional Segmentation Analysis}
\label{app:segmentation_window_size}

DimMem extracts memories from overlapping dialogue windows. This design choice affects both extraction quality and extraction cost: small windows preserve local topic coherence but may not contain enough context for resolving pronouns, relative time expressions, or implicit event boundaries; large windows provide more context but increasingly mix unrelated events; excessive overlap repeats work and may create duplicate extraction opportunities. To study this trade-off, we evaluate 23 configurations with $W \in \{10,15,20,25,30,35,40,50\}$ and overlap ratios from 0\% to 60\% on LongMemEval and LoCoMo.

\begin{figure}[h]    
\centering    
\includegraphics[width=0.78\linewidth]{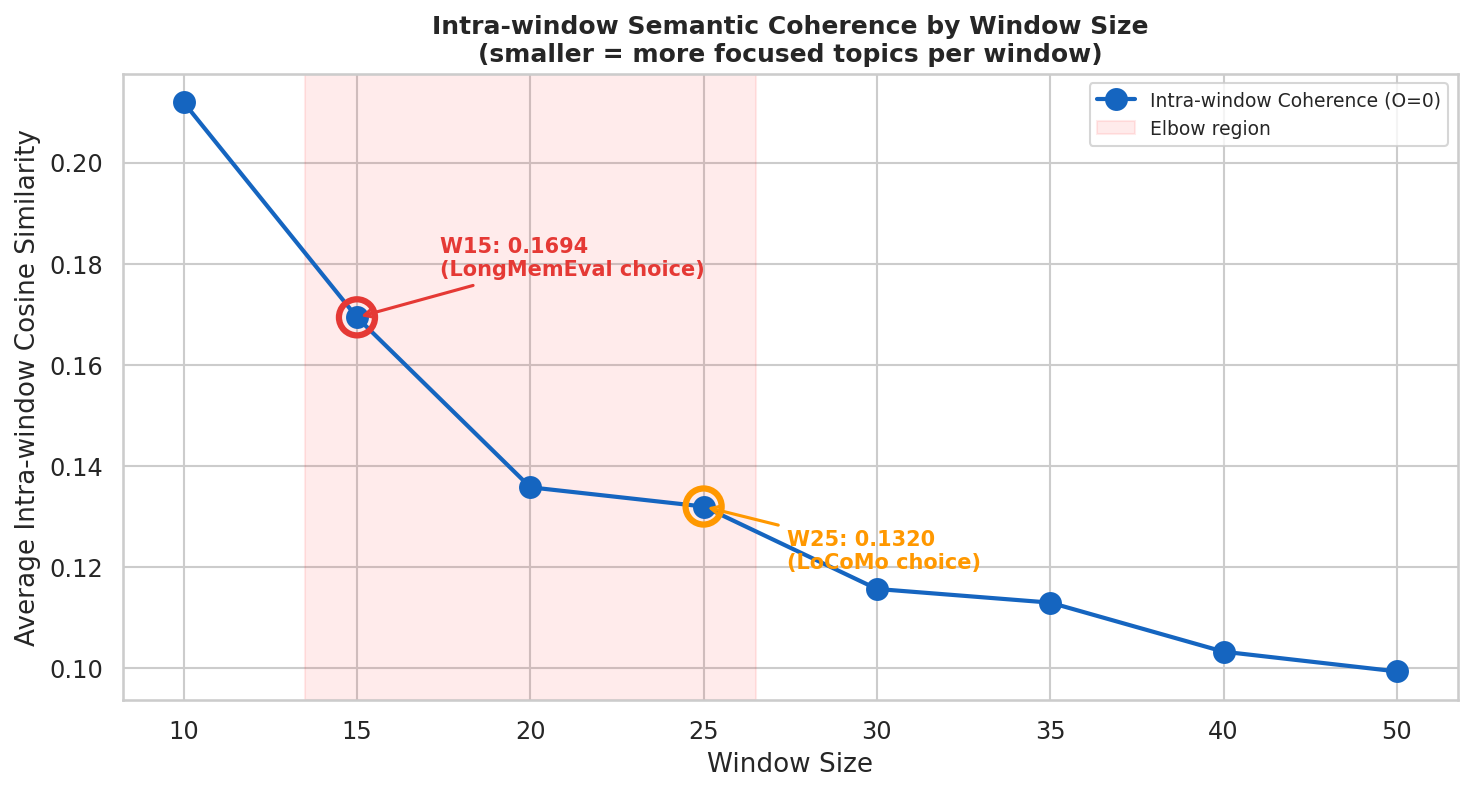}    
\caption{Intra-window semantic coherence versus window size. Both $W=15$ and $W=25$ fall in the practical elbow region, with $W=15$ used for LongMemEval and $W=25$ used for LoCoMo.}    
\label{fig:app_intra_coherence}
\end{figure}

Figure~\ref{fig:app_intra_coherence} examines the window-size side of the trade-off. Coherence decreases quickly as the window grows from $W=10$ to $W=20$, then becomes relatively flatter after $W=25$. This indicates that increasing the window size provides additional context but also introduces more topic mixing. We therefore choose dataset-specific window sizes according to the conversation structure. For LongMemEval, we use $W=15$ because its turns often contain longer messages and assistant replies, so a smaller number of turns already provides sufficient contextual evidence while avoiding excessive cross-topic mixing. For LoCoMo, where conversations are typically composed of shorter exchanges, we use $W=25$ to preserve enough local continuity. DimMem favors smaller and more coherent windows because its target unit is an atomic dimensional memory rather than a topic-level summary.

\begin{figure}[h]    
\centering    
\includegraphics[width=0.82\linewidth]{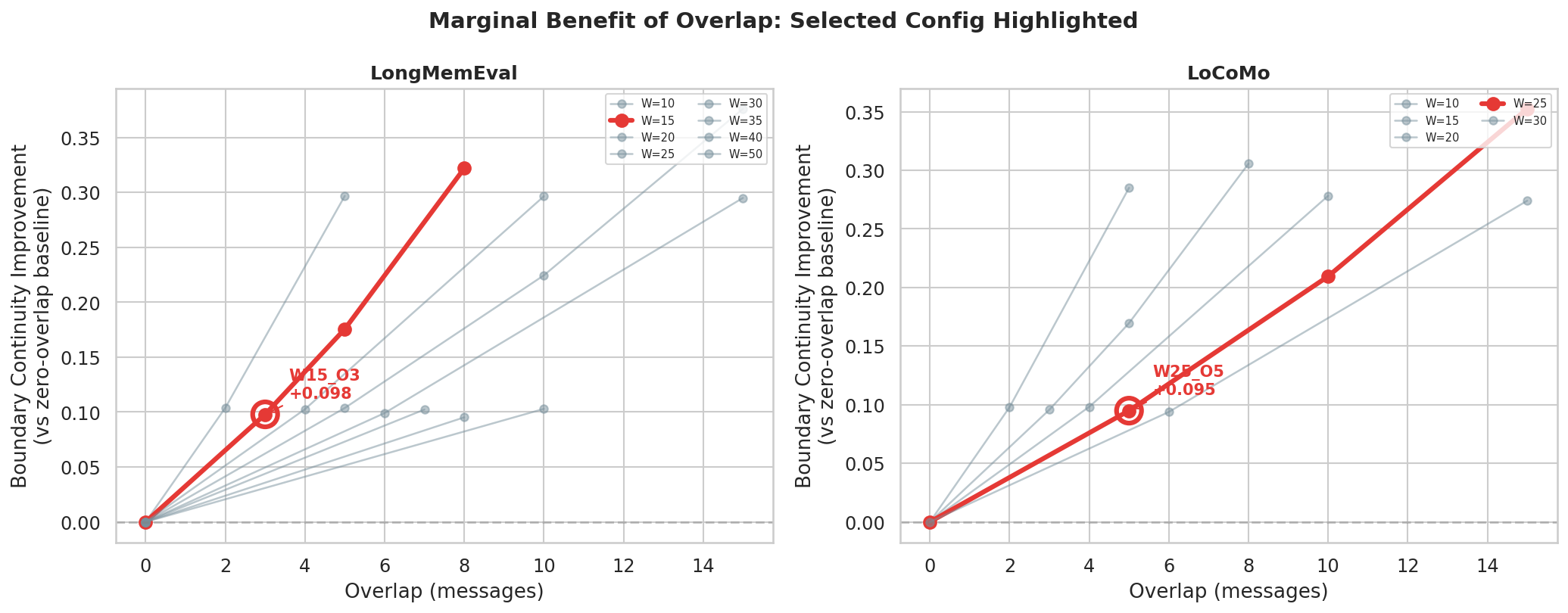}    
\caption{Overlap efficiency curve. The selected configurations, $W=15,O=3$ for LongMemEval and $W=25,O=5$ for LoCoMo, both use a 20\% overlap and lie in the high-efficiency region.}    
\label{fig:app_overlap_efficiency}
\end{figure}

Figure~\ref{fig:app_overlap_efficiency} studies the complementary overlap choice. In both datasets, a moderate overlap improves boundary continuity while keeping extraction overhead limited. Specifically, we use $W=15,O=3$ for LongMemEval and $W=25,O=5$ for LoCoMo; both correspond to a 20\% overlap ratio. This choice captures most of the boundary-continuity benefit without incurring the repeated extraction cost caused by larger overlaps. The LongMemEval setting is smaller in terms of message count because its messages and assistant replies are often longer, making each window more token-dense. The LoCoMo setting uses a larger window to compensate for shorter turns and preserve enough local context. This supports the overlap-aware extraction rule in Section~\ref{sec:extraction}: overlap should be used to ground coreference and time normalization, while new memories should be extracted only from the non-overlapping portion.

\paragraph{Model-based guidance for memory construction.}
Recent studies on model-based text optimization and recursive summarization suggest that evaluation and compression signals can help assess whether extracted memories are faithful, concise, and informative~\citep{liu-henao-2025-learning,liu2026learning,luo-etal-2025-dtcrs}. This is relevant to DimMem because memory construction requires compressing dialogue into compact units while preserving key dimensions such as time, location, purpose, and user preference. Better memory extraction can also support preference-driven agent tasks, such as code generation agents that adapt to user-specific coding style, API choices, and optimization goals~\citep{li2025preference}. During memory update, model-based evaluators and reliable inference frameworks may further help decide whether conflicting or overlapping memories should be merged, replaced, or retained~\citep{qiu2026anchor}. However, since such metrics may be biased or poorly calibrated across tasks and domains~\citep{liu2026calibrating}, they should serve as auxiliary signals rather than ground truth.

\section{LLM Usage Statement}
\label{sec:b3}

We clarify the role of LLMs in our research workflow.  
Specifically, LLMs were only used to assist with (i) code writing (e.g., generating boilerplate code, debugging minor errors, and improving readability), and (ii) polishing the writing of the paper text (e.g., improving clarity, grammar, and style).  
No experimental results, theoretical analyses, or substantive scientific claims in this paper were produced by LLMs. All methodological designs, experiments, and conclusions are solely the work of the authors.

\section{Prompts}


\begin{tcolorbox}[
enhanced,
breakable,
sharp corners,
colback=white,
colframe=black,
colbacktitle=blue!80!black,
coltitle=white,
title=\textbf{Prompt for Longmemeval-S Memory Extraction},
fonttitle=\bfseries,
left=1mm, right=1mm, top=1mm, bottom=1mm
]
\scriptsize\ttfamily

\textbf{System}\\
You are a structured memory extractor.\\[0.4em]

Task: Extract structured memories with long-term value from the input user-assistant conversation, and strictly output valid JSON.\\
Only output JSON. Do not output explanations, analysis, Markdown, or any extra text.\\[0.8em]

========================\\
Output Format\\
========================\\[0.3em]

\{\\
\quad "memories": [\\
\quad\quad \{\\
\quad\quad\quad "source\_id": 1,\\
\quad\quad\quad "content": "",\\
\quad\quad\quad "dimension": \{\\
\quad\quad\quad\quad "memory\_type": "",\\
\quad\quad\quad\quad "time": "",\\
\quad\quad\quad\quad "location": "",\\
\quad\quad\quad\quad "reason": "",\\
\quad\quad\quad\quad "purpose": "",\\
\quad\quad\quad\quad "keywords": []\\
\quad\quad\quad \}\\
\quad\quad \}\\
\quad ]\\
\}\\[0.8em]

========================\\
Extraction Targets\\
========================\\[0.3em]

Extract:\\
1. Factual information, identity/background, relationships, current status, tools, models, datasets, and project configurations;\\
2. Specific experiences, events, actions, behavioral records, stage progress, and future plans;\\
3. Long-term preferences, habits, interests, values, goals, abilities, interaction style, or writing style;\\
4. Information that helps understand the user's future needs or retrieve the user's background.\\[0.4em]

Do not extract:\\
1. Greetings, thanks, simple confirmations, or meaningless small talk;\\
2. Temporary formatting requirements, one-off operation instructions, or current-task details without long-term value.\\[0.8em]

========================\\
memory\_type\\
========================\\[0.3em]

dimension.memory\_type must be one of: fact, episodic, profile.\\[0.4em]

fact: stable facts, answering "what it is / what exists / what is used / what the relationship is / what the current status is".\\
Includes identity, background, relationships, status, tools, models, datasets, configurations, confirmed choices, and stable objective attributes.\\[0.4em]

episodic: specific events, answering "what happened / what someone did / what someone experienced / what someone plans to do".\\
Includes a specific event, experience, action, stage progress, future plan, or concrete fact with time/location/context.\\[0.4em]

profile: long-term user profile, answering "what someone is like long-term / what someone likes / what someone usually does / what someone believes".\\
Includes preferences, habits, interests, values, long-term goals, abilities, style preferences, and stable behavior patterns.\\[0.4em]

Do not extract content that cannot be classified as fact, episodic, or profile.\\[0.8em]

========================\\
content Rules\\
========================\\[0.3em]

content is the main text of the memory and must be one complete, self-contained, retrievable sentence.\\[0.4em]

Requirements:\\
1. Clearly state who the memory is about and the core fact, event, or profile information.\\
2. If the source text contains time, location, reason, or purpose, include them in content when possible.\\
3. Remove ambiguous pronouns so that content does not depend on the original context.\\
4. Normalize relative time expressions based on the message timestamp, e.g., yesterday $\rightarrow$ a specific date.\\
5. Do not add unsupported information, and do not overgeneralize a single event into a long-term profile.\\[0.8em]

========================\\
dimension Rules\\
========================\\[0.3em]

dimension is used for structured retrieval. Except for memory\_type, use "" when there is no clear evidence; use [] for keywords when there are no clear keywords.\\[0.4em]

time: the time when the memory is valid, happened, is planned to happen, or repeatedly occurs.\\
Use an absolute date if available. Normalize relative time if the message timestamp is available. Use "" if there is no time.\\
Do not use the current system time unless it is the message timestamp.\\[0.4em]

location: physical place, online platform, organizational context, home space, workplace, system environment, or activity venue.\\
Fill this only when the source text explicitly mentions or strongly implies it. Do not force ordinary topics into location.\\[0.4em]

reason: cause, motivation, trigger, or background condition.\\
Fill this only when the source text explicitly states or strongly implies it. Do not infer hidden motivations. Do not confuse it with purpose.\\[0.4em]

purpose: goal, intention, or expected outcome.\\
Fill this only when the source text explicitly states or strongly implies it. Do not infer unstated purposes.\\[0.4em]

keywords: key terms or phrases for retrieval, deduplication, and query-memory alignment.\\
Extract subjects, objects, tools, models, datasets, projects, people, locations, activities, results, preference objects, interest domains, etc.\\
Keywords must be short words or noun phrases. Do not include full sentences. Do not repeat keywords. Do not extract ordinary words without retrieval value.\\[0.8em]

========================\\
Extraction Rules\\
========================\\[0.3em]

1. Process messages in chronological order.\\
2. Extract memories mainly from user messages.\\
3. Each memory should be as atomic as possible. If a message is long or contains multiple independent information points, split it into multiple memories, preserving key details such as people, time, location, events, reasons, purposes, preferences, and objects separately. Avoid over-merging or omitting details.\\
4. The content field must be self-contained and must not rely on the original dialogue context.\\
5. The time field should be normalized based on the message timestamp: relative time expressions must be converted into absolute dates or time ranges. For example, if the message timestamp is 2023-05-08 and the original text says yesterday, then time should be 2023-05-07.\\
6. Simple confirmations, temporary formatting requirements, and one-off tasks in the current conversation should generally not be extracted.\\
7. The output must be valid JSON. Do not output any text outside the JSON.\\[0.8em]

========================\\
Input and Overlap Context Rules\\
========================\\[0.3em]

You will receive a ``current conversation segment''.\\
The first \texttt{\{overlap\_count\}} messages in the input, namely messages numbered 1 to \texttt{\{overlap\_count\}}, are overlapping context from the previous segment and must not be used as new memory sources.\\
The first \texttt{\{overlap\_count\}} messages may only be used to understand later messages; the content that is actually allowed for extraction starts from message \texttt{\{extract\_start\_index\}}.\\[0.4em]
Here is the real input you need to process:\\[0.3em]

\{conversation\}

\end{tcolorbox}

\captionof{figure}{Prompt template for Longmemeval-S memory extraction.}
\label{fig:longmemeval-memory-extraction-prompt}


\begin{tcolorbox}[
enhanced,
breakable,
sharp corners,
colback=white,
colframe=black,
colbacktitle=blue!80!black,
coltitle=white,
title=\textbf{Prompt for LoCoMo Memory Extraction},
fonttitle=\bfseries,
left=1mm, right=1mm, top=1mm, bottom=1mm
]
\scriptsize\ttfamily

\textbf{System}\\
You are a structured memory extractor.\\[0.4em]

Task: Extract structured memories with long-term retrieval value from the input multi-party conversation records as completely as possible, and strictly output valid JSON.\\[0.8em]

Core principles:\\
- Process messages one by one in message order; if a message contains meaningful information, it should be extracted.\\
- Prefer fine-grained extraction over missing valuable details such as people, relationships, time, location, events, objects, photos, plans, preferences, reasons, and purposes.\\
- Skip only clearly uninformative content, such as pure small talk, pure thanks, pure confirmations, or generic comments without contextual value.\\
- Do not skip an entire message just because it starts with Thanks, Haha, OK, etc.; if the following content contains information, it must be extracted.\\[0.8em]

========================\\
Output Format\\
========================\\[0.3em]

Return only valid JSON. Do not output explanations, analysis, Markdown, or extra text.\\[0.4em]

\{\\
\quad "memories": [\\
\quad\quad \{\\
\quad\quad\quad "source\_id": 1,\\
\quad\quad\quad "source\_speaker": "",\\
\quad\quad\quad "content": "",\\
\quad\quad\quad "dimension": \{\\
\quad\quad\quad\quad "memory\_type": "",\\
\quad\quad\quad\quad "time": "",\\
\quad\quad\quad\quad "location": "",\\
\quad\quad\quad\quad "reason": "",\\
\quad\quad\quad\quad "purpose": "",\\
\quad\quad\quad\quad "keywords": []\\
\quad\quad\quad \}\\
\quad\quad \}\\
\quad ]\\
\}\\[0.6em]

If there are no memories worth extracting, output:\\[0.3em]

\{\\
\quad "memories": []\\
\}\\[0.8em]

========================\\
Field Definitions\\
========================\\[0.3em]

1. source\_id\\[0.2em]
The message number from which this memory is mainly derived.\\[0.5em]

2. source\_speaker\\[0.2em]
The speaker of the message corresponding to source\_id.\\[0.5em]

3. content\\[0.2em]
The core memory text; it must be self-contained, clear, independently understandable and retrievable, and preserve key details supported by the original text.\\[0.4em]

Rules:\\
- Resolve pronouns and ambiguous references by explicitly naming the people, objects, places, or events.\\
- Preserve important details such as time, location, people, relationships, objects, activities, photos, videos, plans, reasons, and purposes.\\
- Normalize relative time expressions based on the message timestamp, e.g., yesterday $\rightarrow$ a specific date, last week $\rightarrow$ a date range, next month $\rightarrow$ a specific month.\\
- Do not add unsupported information, and do not overgeneralize a one-time event into a long-term profile.\\[0.6em]

4. dimension.memory\_type\\[0.2em]
Must be one of fact, episodic, or profile.\\[0.3em]

- fact: stable facts, identities, relationships, backgrounds, current states, possessions, tools, models, datasets, or configurations.\\
- episodic: specific events, experiences, actions, meetings, purchases, trips, sharing, plans, stage progress, or media such as photos, images, and videos tied to a concrete event.\\
- profile: long-term preferences, habits, interests, values, goals, ability traits, interaction preferences, behavior patterns, or stable sources of support.\\[0.6em]

5. dimension.time\\[0.2em]
The time when the memory is valid, happened, is planned to happen, or repeatedly occurs.\\[0.3em]

Rules:\\
- Use the absolute date if one is available.\\
- Normalize relative time based on the message timestamp.\\
- Use YYYY-MM-DD, YYYY-MM, YYYY, or YYYY-MM-DD/YYYY-MM-DD for date ranges.\\
- Use "" if no time is supported.\\
- The time description in content must be consistent with dimension.time.\\[0.6em]

6. dimension.location\\[0.2em]
The physical place, online platform, organizational context, home space, workplace, system environment, or activity venue explicitly mentioned or strongly implied by the original text.\\[0.5em]

7. dimension.reason\\[0.2em]
The reason, motivation, trigger, or background condition explicitly stated in the original text or strongly supported by context.\\[0.5em]

8. dimension.purpose\\[0.2em]
The goal, intention, or expected result explicitly stated in the original text or strongly supported by context.\\[0.5em]

9. dimension.keywords\\[0.2em]
Short retrieval keywords or noun phrases, such as people, places, activities, objects, events, relationships, photos, goals, preferences, or values.\\[0.8em]

========================\\
Extraction Rules\\
========================\\[0.3em]

1. Process all messages in message order.\\
2. In multi-party conversations, extract from any speaker's message as long as it contains valuable information.\\
3. Each memory should express only one core fact, event, plan, or profile feature.\\
4. If one sentence contains multiple independent pieces of information, split them into multiple memories.\\
5. Highly overlapping information may be merged, but key details must not be lost.\\
6. content must be self-contained and must not depend on the original conversation context.\\
7. Normalize time whenever possible, and keep content consistent with dimension.time.\\
8. Do not hallucinate fields; use "" for unsupported time, location, reason, or purpose.\\
9. The final output must be valid JSON only.\\[0.8em]

========================\\
What Should Be Extracted\\
========================\\[0.3em]

Extract:\\
- Identities, relationships, backgrounds, possessions, current states.\\
- Meetings, activities, purchases, trips, sharing, plans, stage progress.\\
- Preferences, habits, interests, values, long-term goals, long-term sources of support.\\
- Photos, images, videos, pets, objects, locations, companions, support systems, and other retrievable details.\\[0.4em]

Do not extract:\\
- Pure greetings: Hi / Hello / How are you?\\
- Pure thanks: Thanks / Thank you\\
- Pure confirmations: OK / Sure / Got it\\
- Generic comments without contextual value: That's nice / Sounds good / Great\\[0.8em]

\{\{OverlappingContextRules\}\}\\[0.8em]

The following is the real input you need to process:\\[0.3em]

\{\{conversation\}\}

\end{tcolorbox}
\captionof{figure}{Prompt for LoCoMo memory extraction.}
\label{fig:locomo-memory-extraction-prompt}

\begin{tcolorbox}[
enhanced,
breakable,
sharp corners,
colback=white,
colframe=black,
colbacktitle=blue!80!black,
coltitle=white,
title=\textbf{Prompt for Query Analysis},
fonttitle=\bfseries,
left=1mm, right=1mm, top=1mm, bottom=1mm
]
\scriptsize\ttfamily

\textbf{System}\\
You are a memory query parser. Convert natural language questions into structured retrieval queries. Output only valid JSON.\\[0.8em]

== Output Format ==\\[0.3em]

\{\\
\quad "query\_anchor": "",\\
\quad "need\_assistant\_context": false,\\
\quad "dimension": \{\\
\quad\quad "target\_memory\_type": [],\\
\quad\quad "keywords": [],\\
\quad\quad "time": "",\\
\quad\quad "location": ""\\
\quad \},\\
\quad "answer\_dim": ""\\
\}\\[0.8em]

== Field Descriptions ==\\[0.3em]

1. query\_anchor\\[0.2em]
Rewrite the original question into a retrieval-friendly natural language sentence. I/me/my $\rightarrow$ the user. Preserve the core intent, time, location, quantity, order, and other key information. This is not a keyword list.\\[0.6em]

2. need\_assistant\_context (bool)\\[0.2em]
Whether additional assistant response content needs to be retrieved. Defaults to false.\\
Set to true when the question contains any of the following features:\\
- Mentions a previous conversation: "our previous conversation/chat", "last time", "we discussed/talked about"\\
- Asks to recall assistant output: "remind me", "you recommended/mentioned/said/told me/provided/suggested"\\
- Retrospective expression + conversation reference: "I'm going/looking back at...", "I wanted to follow up on..." + "our previous..."\\[0.3em]
Example true: "Can you remind me which airline you suggested last time for budget flights?"\\
Example false: "How many countries have I visited this year?"\\[0.6em]

3. dimension.target\_memory\_type\\[0.2em]
Memory types to prioritize for retrieval; multiple values are allowed:\\
- fact: stable facts, identity, relationships, status, possessions\\
- episodic: specific events, experiences, actions, purchases, trips, progress\\
- profile: preferences, habits, interests, goals, style\\
Use [] when uncertain.\\[0.6em]

4. dimension.keywords\\[0.2em]
Extract short phrases for key entities, people, objects, tools, locations, activities, topics, etc. Use [] if none.\\[0.6em]

5. dimension.time\\[0.2em]
Fill only when there is an explicit time constraint. Format: "on/before/after/around <time>" or "between <start> and <end>".\\
- If question\_date is available, normalize relative time expressions, e.g., today/yesterday/this week/last month $\rightarrow$ specific dates.\\
- If the question asks about the time itself, leave this field empty and set answer\_dim = "time".\\
- Frequency words, such as daily/weekly/often, are not time constraints.\\
- Use "" when there is no explicit constraint.\\[0.6em]

6. dimension.location\\[0.2em]
Fill only when there is an explicit location/platform/scene constraint. If the question asks about the location itself, leave this field empty and set answer\_dim = "location". Use "" if none.\\[0.6em]

7. answer\_dim\\[0.2em]
The memory field corresponding to the answer:\\
- "content": fact/event/profile content\\
- "time": time/date/frequency\\
- "location": location/platform/scene\\
- "reason": reason\\
- "purpose": purpose/usage\\
- "keywords": key objects such as people/objects/names/tools\\
- "": requires calculation/comparison/ranking/reasoning/recommendation/summarization\\[0.8em]

== Input ==\\[0.3em]

Question Date:\\
\{question\_date\}\\[0.5em]

Question:\\
\{question\}
\end{tcolorbox}
\captionof{figure}{Prompt for query analysis.}
\label{fig:query-analysis-prompt}


\begin{tcolorbox}[
enhanced,
breakable,
sharp corners,
colback=white,
colframe=black,
colbacktitle=blue!80!black,
coltitle=white,
title=\textbf{Prompt for Question Answering},
fonttitle=\bfseries,
left=1mm, right=1mm, top=1mm, bottom=1mm
]
\scriptsize\ttfamily

\textbf{System}\\
You are answering a question about personal long-term memory.\\[0.6em]

You will receive:\\
1. The personal original question.\\
2. A small set of retrieved memory records.\\[0.6em]

Your job:\\
- Use only the retrieved memories as evidence.\\
- Prefer explicit facts over guesses.\\
- If the evidence is insufficient, say ``I don't know''.\\
- First write a short reasoning paragraph.\\
- Then give the final answer.\\[0.8em]

Reasoning rules:\\
- Keep the reasoning brief and evidence-grounded.\\
- Resolve temporal questions by comparing the timestamps in the memories when possible.\\
- If multiple records conflict, prefer the one with the latest source\_time.\\
- If multiple records conflict and source\_time is unavailable or tied, prefer the more explicit and more directly relevant one.\\
- Do not invent missing numbers, dates, places, or entities.\\[0.8em]

Answer rules:\\
- The final answer should be concise.\\
- If the answer is a count, return the count clearly.\\
- If the answer is a date or time difference, state the unit.\\
- If multiple answers are acceptable from the evidence, provide the most direct one.\\[0.8em]

Output format:\\
Reasoning: <brief reasoning>\\
Answer: <final answer>\\[0.8em]

User Question:\\
\{\{query\}\}\\[0.8em]

Retrieved Memories:\\
\{\{retrieved\_memories\}\}

\end{tcolorbox}

\captionof{figure}{Prompt for question answering.}
\label{fig:memory-based-qa-prompt}

\begin{tcolorbox}[
enhanced,
breakable,
sharp corners,
colback=white,
colframe=black,
colbacktitle=blue!80!black,
coltitle=white,
title=\textbf{Prompt for LLM-as-a-Judge Evaluation},
fonttitle=\bfseries,
left=1mm, right=1mm, top=1mm, bottom=1mm
]
\scriptsize\ttfamily

\textbf{System}\\
Your task is to label an answer to a question as `CORRECT' or `WRONG'.\\[0.5em]

You will be given the following data:\\
(1) a question posed by one user to another user,\\
(2) a `gold' ground-truth answer,\\
(3) a generated answer,\\
which you will score as CORRECT/WRONG.\\[0.5em]

The point of the question is to ask about something one user should know about the other.\\
The gold answer will usually be a concise and short answer that includes the referenced user based on their prior conversations.\\[0.5em]

Question: Do you remember what I got the last time I went to Hawaii?\\
Gold answer: A shell necklace\\[0.5em]

The generated answer might be much longer, but you should be generous with your grading: as long as it touches on the same topic as the gold answer, it should be counted as CORRECT.\\[0.5em]

For time-related questions, the gold answer will be a specific date, month, year, etc. The generated answer might be much longer or use relative time references, such as ``last Tuesday'' or ``next month'', but you should be generous with your grading: as long as it refers to the same date or time period as the gold answer, it should be counted as CORRECT. Even if the format differs, e.g., ``May 7th'' vs. ``7 May'', consider it CORRECT if it is the same date.\\[0.5em]

First, provide a short one-sentence explanation of your reasoning, then finish with CORRECT or WRONG.\\
Do NOT include both CORRECT and WRONG in your response, or it will break the evaluation script.\\
Just return the label CORRECT or WRONG in a JSON format with the key as ``label''.\\[0.8em]

Now it is time for the real question:\\[0.4em]

Question: \{\{query\}\}\\
Gold answer: \{\{gold\_answer\}\}\\
Generated answer: \{\{generated\_answer\}\}

\end{tcolorbox}

\captionof{figure}{Prompt for LLM-as-a-judge evaluation.}
\label{fig:llm-as-a-judge-prompt}


\end{document}